\documentclass[runningheads]{llncs}
\usepackage{graphicx}

\usepackage{tikz}
\usepackage{comment}
\usepackage{amsmath,amssymb} 
\usepackage{color}
\usepackage{float}
\usepackage{colortbl} 
\usepackage{xcolor} 
\usepackage{cite}
\definecolor{MyBlue}{rgb}{0.25,0.5,0.75}
\colorlet{shade1}{MyBlue!80}
\colorlet{shade2}{MyBlue!65}
\colorlet{shade3}{MyBlue!50}
\colorlet{shade4}{MyBlue!45}
\colorlet{shade5}{MyBlue!20}
\colorlet{shade6}{MyBlue!10}

\usepackage{caption}
\usepackage{subcaption}

\usepackage[accsupp]{axessibility}  

\usepackage{etoolbox}
\patchcmd{\paragraph}{\itshape}{\bfseries\boldmath}{}{}

\usepackage{xspace}
\newcommand*{\eg}{e.g.\@\xspace}
\newcommand*{\ie}{i.e.\@\xspace}
\newcommand*{\cf}{c.f.\@\xspace}

\usepackage[pagebackref,breaklinks,colorlinks]{hyperref}

\begin{document}
\pagestyle{headings}
\mainmatter
\def\ECCVSubNumber{24}  

\title{Neural Mesh-Based Graphics}

%

\author{\large Shubhendu Jena, Franck Multon, Adnane Boukhayma}
\authorrunning{S. Jena et al.}
%
\institute{\normalsize Inria, Univ. Rennes, CNRS, IRISA, M2S, France}
\maketitle

\begin{abstract}
We revisit NPBG \cite{aliev2020neural}, the popular approach to novel view synthesis that introduced the ubiquitous point feature neural rendering paradigm. We are interested in particular in data-efficient learning with fast view synthesis. We achieve this through a view-dependent mesh-based denser point descriptor rasterization, in addition to a foreground/background scene rendering split, and an improved loss. By training solely on a single scene, we outperform NPBG \cite{aliev2020neural}, which has been trained on ScanNet \cite{dai2017scannet} and then scene finetuned. We also perform competitively with respect to the state-of-the-art method SVS~\cite{riegler2021stable}, which has been trained on the full dataset (DTU\cite{aanaes2016large} and Tanks and Temples~\cite{knapitsch2017tanks}) and then scene finetuned, in spite of their deeper neural renderer.
\end{abstract}

\section{Introduction}

Enabling machines to understand and reason about 3D shape and appearance is a long standing goal of computer vision and machine learning, with implications in numerous 3D vision downstream tasks. In this respect, novel view synthesis is a prominent computer vision and graphics problem with rising applications in free viewpoint, virtual reality, image editing and manipulation, as well as being a corner stone of building an efficient metaverse. The introduction of deep learning in the area of novel view synthesis brought higher robustness and  generalization in comparison to earlier traditional approaches. While the current trend is learning neural radiance fields (\eg \cite{mildenhall2020nerf, yariv2021volume, wang2021neus}), training and rendering such implicit volumetric models still presents computational challenges, despite recent efforts towards alleviating these burdens (\eg \cite{hedman2021baking, yu2021plenoctrees, sitzmann2021light,li2022learning}).
An appealing alternate learning strategy \cite{aliev2020neural, riegler2021stable, riegler2020free, thies2019deferred}, achieving to date state-of-the-art results on large outdoors unbounded scenes such as the Tanks and Temples dataset \cite{knapitsch2017tanks}, consists of using a pre-computed geometric representation of the scene to guide the novel view synthesis process. As contemporary successors to the original depth warping techniques, these methods benefit from a strong geometry prior to constrain the learning problem, and recast its 3D complexity into a simpler 2D neural rendering task, providing concurrently faster feed-forward inference.

Among the latter, NPBG~\cite{aliev2020neural} is a popular strategy, being core to several other neural rendering based methods (\eg \cite{wu2020multi, prokudin2021smplpix, zakharkin2021point, Raj_2021_CVPR}). Learnable descriptors are appended to the geometry points, and synthesis consists of rasterizing then neural rendering these features. It is practical also as it uses a lightweight and relatively simpler architecture, compared to competing methods (\eg \cite{riegler2020free, riegler2021stable}).

Our motivation is seeking a data-efficient, fast, and relatively lightweight novel-view synthesis method. Hence we build on the idea of NPBG~\cite{aliev2020neural}, and we introduce several improvements allowing it to scale to our aforementioned goals. In particular, and differently from NPBG~\cite{aliev2020neural}: We introduce denser rasterized feature maps through a combination of denser point rasterization and face rasterization; We enforce view-dependency explicitly through anisotropic point descriptors; As the foreground and background geometries differ noticeably in quality, we propose to process these two feature domains separately to accommodate independently for their respective properties; Finally, we explore a self-supervised loss promoting photo-realism and generalization outside the training view corpus. The improvement brought by each of these components is showcased in Table \ref{tab:table3}.

By training simply on a single scene, our method outperforms NPBG~\cite{aliev2020neural}, even though it has been additionally trained on ScanNet~\cite{dai2017scannet} and then further fine-tuned on the same scene, in terms of PSNR, SSIM~\cite{wang2004image} and LPIPS~\cite{zhang2018unreasonable}, and both on the Tanks and Temples~\cite{knapitsch2017tanks} (Tab. \ref{tab:table1}) and DTU~\cite{aanaes2016large} (Tab. \ref{tab:table4}) datasets. The performance gap is considerably larger on DTU~\cite{aanaes2016large}.
Our data efficiency is also illustrated in the comparison to state-of-the-art method SVS \cite{riegler2021stable} (Tab. \ref{tab:table1} \ref{tab:table4}). We achieve competitive numbers despite their full dataset trainings, their deeper convolutional network based neural renderers, and slower inference. We also recover from some of their common visual artifacts as shown in Figure \ref{fig:qualitative_T_T}.  


\section{Related Work}

While there is a substantial body of work on the subject of novel view synthesis, we review here work we deemed most relevant to the context of our contribution.\\ 
\\{\bf Novel view synthesis.} The task of novel view synthesis essentially involves using observed images of a scene and the corresponding camera parameters to render images from novel unobserved viewpoints. This is a long explored problem, with early non deep-learning based methods~\cite{chen1993view, seitz1996view, zitnick2004high, debevec1996modeling, gortler1996lumigraph, levoy1996light} using a set of input images to synthesize new views of a scene. However, these methods impose restrictions on the configuration of input views. For example,~\cite{levoy1996light} requires a dense and regularly spaced camera grid while~\cite{davis2012unstructured} requires that the cameras are located approximately on the surface of a sphere and the object is centered. To deal with these restrictions, unstructured view synthesis methods use a 3D proxy geometry of the scene to guide the view synthesis process~\cite{buehler2001unstructured, kopf2014first}. With the rise of deep-learning, it has also come to be used extensively for view synthesis, either by blending input images to synthesize the target views~\cite{hedman2018deep, thies2018ignor}, or by learning implicit neural radiance fields followed by volumetric rendering to generate the target views~\cite{mildenhall2020nerf}, or even by using a 3D proxy geometry representation of the scene to construct neural scene representations~\cite{thies2019deferred, niemeyer2020differentiable, aliev2020neural, riegler2020free, riegler2021stable}.\\
\\{\bf View synthesis w/o geometric proxies.} Early deep-learning based approaches combine warped or unwarped input images by predicting the corresponding blending weights to compose the target view~\cite{thies2018ignor, choi2019extreme}. Thereafter, several approaches came up leveraging different avenues such as predicting camera poses~\cite{zhou2017unsupervised}, depth maps~\cite{kalantari2016learning}, multi-plane images~\cite{flynn2016deepstereo, zhou2018stereo}, and voxel grids~\cite{kar2017learning}. Recently, implicit neural radiance fields (NeRF)~\cite{mildenhall2020nerf} has emerged as a powerful representation for novel view synthesis. It uses MLPs to map spatial points to volume density and view-dependent colors. Hierarchical volumetric rendering is then performed on the predicted point colors to render the target image. Some of the problems associated with NeRF~\cite{mildenhall2020nerf} include higher computational cost and time of rendering complexity, the requirement of dense training views, the lack of across-scene generalization, and the need for test-time optimization. A number of works~\cite{jain2021putting, niemeyer2021regnerf, barron2021mip, liu2020neural, yu2021pixelnerf, yu2021plenoctrees, garbin2021fastnerf, reiser2021kilonerf, wizadwongsa2021nex} have tried addressing these limitations. In particular, Spherical Harmonics~\cite{sloan2002precomputed} have been used to speed up inference of Neural Radiance Fields by factorizing the view-dependent appearance~\cite{yu2021plenoctrees}. Nex~\cite{wizadwongsa2021nex} introduced a related idea, where several basis functions such as Hemi-spherical harmonics, Fourier Series, and Jacobi Spherical Harmonics were investigated, and concluded that learnable basis functions offer the best results. Some other methods, like pixelNeRF~\cite{yu2021pixelnerf}, GRF~\cite{trevithick2021grf}, IBRNet~\cite{wang2021ibrnet} and MVSNeRF~\cite{chen2021mvsnerf} proposed to augment NeRFs~\cite{mildenhall2020nerf} with 2D and 3D convolutional features collected from the input images. Hence, they offer forward pass prediction models, i.e. test-time optimization free, while introducing generalization across scenes. While these methods are promising, they need to train on full datasets to generalize well, while at the same time evaluating hundreds of 3D query points per ray for inference similar to NeRF~\cite{mildenhall2020nerf}. Hence, both training and inference often takes quite long for these methods. We note that besides encoders \cite{yu2021pixelnerf,trevithick2021grf,wang2021ibrnet,chen2021mvsnerf}, implicit neural representations can also be conditioned through meta-learning \eg \cite{ouasfi2022few,sitzmann2020metasdf,sitzmann2019scene}.\\
\\{\bf View synthesis using geometric proxies.}
Different from the work we have discussed so far, several recent methods utilize a geometric reconstruction of the scene to construct neural scene representations and consequently use them to synthesize target views. These geometric proxies can either be meshes~\cite{riegler2020free, riegler2021stable, thies2019deferred} or point clouds~\cite{aliev2020neural, meshry2019neural, pittaluga2019revealing, song2020deep}. SVS~\cite{riegler2021stable} and FVS~\cite{riegler2020free} utilize COLMAP~\cite{schonberger2016structure, schonberger2016pixelwise} to construct a mesh scaffold which is then used to select input views with maximum overlap with the target view. FVS~\cite{riegler2020free} then uses a sequential network based on gated recurrent units (GRUs) to blend the encoded input images. SVS~\cite{riegler2021stable} operates on a similar principle except that the geometric mesh scaffold is used for on-surface feature aggregation, which entails processing or aggregating directional feature vectors from encoded input images at each 3D point to produce a new feature vector for a ray that maps this point into the new target view. Deferred neural rendering (DNR)~\cite{thies2019deferred} proposes to learn neural textures encoding the point plenoptic function at different surface points alongside a neural rendering convolutional network. It is infeasible for very large meshes due to the requirement of UV-parametrization. On the other hand, NPBG~\cite{aliev2020neural} operates on a similar idea by learning neural descriptors of surface elements jointly with the rendering network, but uses point-based geometry representation instead of meshes. Recently, NPBG++~\cite{rakhimov2022npbg++}, which is a concurrent work, extended NPBG~\cite{aliev2020neural} by using 2D convolutional features from source images and learnable basis functions to model view-dependent neural point descriptors. Similarly,~\cite{meshry2019neural, song2020deep} use COLMAP~\cite{schonberger2016structure, schonberger2016pixelwise} to reconstruct a colored point cloud of the scene in question, which is used alongside a neural renderer to render target images. Our method combines both approaches in the sense that it uses both point cloud and mesh representation of the scene as geometric proxies. Like NPBG~\cite{aliev2020neural}, we also learn neural descriptors of surface elements, but since point clouds of large unbounded scenes are often sparse, using meshes helps in enhancing the richness of the rasterized neural descriptors. Hence, using both point clouds and meshes help us in achieving a balance between accuracy and density of the rasterized scene neural descriptors, which we will explain in detail in the upcoming sections and also through an ablative analysis. 

\begin{figure}[t!]
\begin{center}
\includegraphics[width=0.8\linewidth]{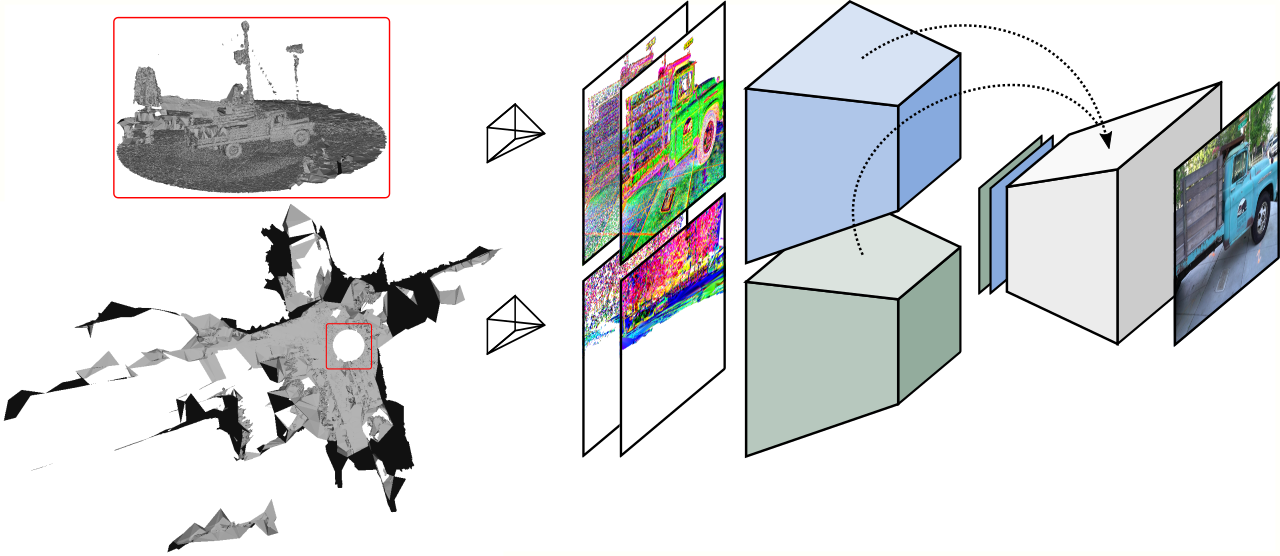}
\end{center}
   \caption{Overview: An automatic split of the scene geometry is used to rasterize foreground/background mesh-borne view-dependent features, through both point based and mesh based rasterizations. A convolutional U-Net~\cite{ronneberger2015u} maps the feature images into the target image.}
\label{fig:pipeline}
\end{figure}

\section{Method}\label{sec:Method}

Given a set of calibrated images of a scene, our goal is to  generate novel views from the same scene through deep learning. To this end, we design a forward pass prediction method, expected to generalize to target views including and beyond input view interpolation. Using a geometry proxy of the scene, we set view-dependent learnable descriptors on the vertices, and we split the scene automatically into a dense foreground and a sparser background. Each of these areas are rasterized through PyTorch3D's point based and mesh based rasterizations~\cite{ravi2020accelerating}. A convolutional neural renderer translates and combines the resulting image features into the target color image. Figure \ref{fig:pipeline} illustrates this pipeline. Our method can be trained on a single scene by fitting the point descriptors and learning the neural renderer weights jointly. It can also benefit from multi-scene training through the mutualization of the neural renderer learning.  
We present in the remaining the different components of our method.

\paragraph{Preprocessing} We need to obtain a geometry representing the scene from the training images as a preprocessing stage. Standard structure-from-motion (SfM) and multi-view stereo (MVS) pipelines can be used to achieve this ~\cite{schonberger2016structure, schonberger2016pixelwise}. In this respect, we chose to use the preprocessed data from SVS~\cite{riegler2021stable} and FVS~\cite{riegler2020free}, and so the preprocessing steps are identical to these methods. The first step involves running structure-from-motion~\cite{schonberger2016structure, schonberger2016pixelwise} on the images to get camera intrinsics $\{K\}$ and camera poses as rotation matrices $\{R\}$, and translations $\{T\}$. The second step involves running multi-view stereo on the posed images, to obtain per-image depth maps, and then fusing these into a point cloud. Finally, Delaunay-based triangulation is applied to the obtained point cloud to get a 3D surface mesh $\mathcal{M}$. These steps are implemented following COLMAP~\cite{schonberger2016structure, schonberger2016pixelwise}.

\begin{figure}[t!]
\begin{center}
\includegraphics[width=0.8\linewidth]{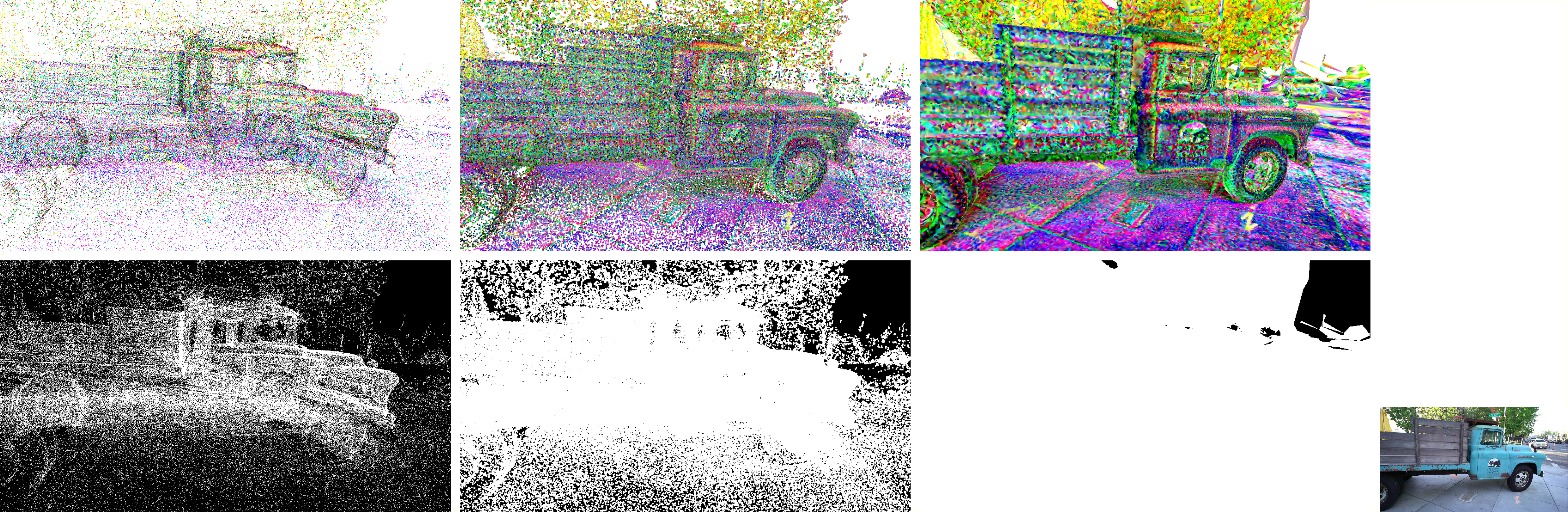}
\end{center}
   \caption{We introduce denser feature images compared to NPBG~\cite{aliev2020neural}. Left: NPBG's rasterization. Center: Our point based rasterization. Right: Our mesh based rasterization. Top row shows feature PCA coefficients. Bottom row shows the resulting rasterization mask (Occupied/unoccupied pixels).}
\label{fig:raster}
\end{figure}

\subsection{Dual Feature Rasterization}\label{sec:FeatureRasterization}

Given a target camera pose $R \in SO(3)$, $T \in \mathbb{R}^3$, the rendering procedure starts with the rasterization of learnable mesh features into the target image domain. The mesh $\mathcal{M}$ consists here of vertices (\ie points) $\mathcal{P} = \{p_1, p_2, ..., p_N\}$ with neural point descriptors $\mathcal{K} = \{k_1, k_2, ..., k_N\}$, and faces $\mathcal{F} = \{f_1, f_2, ..., f_M\}$. 

We noticed initially that having denser feature images improves the performance of the neural rendering. Hence, differently from NPBG~\cite{aliev2020neural}, we use the PyTorch3D~\cite{ravi2020accelerating} renderer which allows us to obtain denser feature images for point cloud based rasterization. Furthermore, we notice additionally that using PyTorch3D's mesh based rasterization provides even denser feature images (\cf Fig.\ref{fig:raster}). Hence, we propose to rasterize the scene geometry features using both modes.

\paragraph{Point Cloud based Rasterization}

PyTorch3D~\cite{ravi2020accelerating} requires us setting a radius $r$ in a normalized space which thresholds the distance between the target view pixel positions $\{(u,v)\}$ and the projected 3D points of the scene onto the target view, \ie $\{\Pi(p):p\in P\}$. If this distance is below the threshold  for a given pixel $(u,v)$, we consider this point to be a candidate for that pixel, and we finally pick the point $p_{u,v}$ with the smallest $z$ coordinate in the camera coordinate frame. This writes:
\begin{gather}
\mathcal{P}_{u,v} = \{p\in \mathcal{P}: ||\Pi(p)-(u,v)||_2 \leq r\}.\\
p_{u,v} = \operatorname*{argmin}_{p\in \mathcal{P}_{u,v}} p_z,\quad \text{where}\quad p = (p_x,p_y,p_z).
\end{gather}
The neural descriptor of the chosen point $p_{u,v}$ is projected onto the corresponding pixel position to construct the rasterized feature image. We set a radius of $r = 0.006$ to achieve a balance between accuracy of the projected point positions and the density of the rasterized feature images. The feature descriptors for each pixel are weighed inversely with respect to the distance of the projected 3D point to the target pixel, which can be expressed as $w_{u,v} = (1 - ||\Pi(p_{u,v})-(u,v)||_2^2)/r^2$ where  $||\Pi(p_{u,v})-(u,v)||_2 < r$. Finally the point feature image $\{k_{u,v}^{\text{pt}}\}$ can hence be expressed as follows: 
\begin{equation}
k_{u,v}^{\text{pt}} = w_{u,v} \times \mathcal{K}(p_{u,v}),
\label{equ:pt_ft}
\end{equation}
where $\mathcal{K}(p_{u,v})$ is the neural descriptor of mesh vertex $p_{u,v}$. 

\paragraph{Mesh based Rasterization}
On the other hand, the mesh rasterizer in PyTorch3D~\cite{ravi2020accelerating} finds the faces of the mesh intersecting each pixel ray and chooses the face with the nearest point of intersection in terms of the z-coordinate in camera coordinate space, which writes:
\begin{gather}
\mathcal{F}_{u,v} = \{f\in \mathcal{F}: (u,v) \in \Pi(f)\}.\\
f_{u,v} = \operatorname*{argmin}_{f\in \mathcal{F}_{u,v}} \hat{f}_z,\quad \text{where}\quad \hat{f} = (\hat{f}_x,\hat{f}_y,\hat{f}_z).
\end{gather}
$\hat{f}$ represents here the intersection between ray $(u,v)$ and face $f$ in camera coordinate frame. 
Finally, to find the feature corresponding to each pixel, the barycentric coordinates of the point of intersection $\hat{f}_{u,v}$ of the face $f_{u,v}$ with the corresponding pixel ray are used to interpolate the neural point descriptors over the face. By noting $(p_i,p_j,p_k)$ as the vertices making up face $f_{u,v}$, the mesh feature image $\{k_{u,v}^{\text{mesh}}\}$ can be expressed as follows:

\begin{equation}
k_{u,v}^{\text{mesh}} = \alpha \mathcal{K}(p_i) +  \beta \mathcal{K}(p_j) + \gamma \mathcal{K}(p_k),
\label{equ:ms_ft}
\end{equation}

where $(\alpha,\beta,\gamma)$ are the barycentric coordinates of $\hat{f}_{u,v}$ in $f_{u,v}$.

While we expect the point cloud rasterization to be more accurate, the mesh rasterization provides a denser feature image, albeit less accurate and fuzzier due to the dependence on the quality of the geometric triangulation being used. The final feature image combines the best of both worlds as it consists of the mesh and point feature rasterized images concatenated together, i.e. $k_{u,v} = \left[k_{u,v}^{\text{pt}},k_{u,v}^{\text{mesh}}\right]$.

\begin{figure}[t!]
\begin{center}
\includegraphics[width=0.9\linewidth]{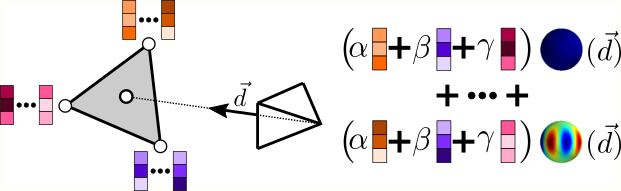}
\end{center}
   \caption{Differently from NPBG~\cite{aliev2020neural}, we introduce mesh-based view-dependent feature rasterization. We learn per-point Spherical Harmonic coefficients, interpolated via the barycentric coordinates of the ray-face intersection.}
\label{fig:ani_feat}
\end{figure}

\subsection{Anisotropic Features}\label{sec:DirnFeatures}

Novel view synthesis entails learning scene radiance functions as perceived appearance can vary according to the viewing angle. While the neural point descriptors in NPBG~\cite{aliev2020neural} are not view direction dependent per se, the neural rendering convolutional network could in theory model such view-dependency, even without taking the camera parameters of the target view explicitly as input. That is, the spatial disposition and neighborhood of these rasterized descriptors in image domain depends on the target view. 
However, incorporating view dependency in the geometry descriptors by design is bound to improve this aspect within such novel view synthesis strategy. Hence, we define anisotropic neural point descriptors in this work, and we implement this idea efficiently using Spherical Harmonics (SH). Spherical Harmonics have been long used as a popular low-dimensional representation for spherical functions to model \eg Lambertian surfaces~\cite{ramamoorthi2001relationship, basri2003lambertian} or even glossy surfaces~\cite{sloan2002precomputed}. They have been also recently introduced as a means of alleviating the computational complexity of volumetric rendering of implicit neural radiance fields (NeRFs) \cite{yu2021plenoctrees}. 

We adapt our point descriptors to auto-decode Spherical Harmonic coefficients rather than point features directly. Reformulating our earlier definition, a descriptor for a point (\ie vertex) $p$ can be expressed now as the set of coefficients:
\begin{equation}
\mathcal{K}(p) = \left(k_p^{l,m}\right)_{0 \leq l \leq l_{max}}^{-l \leq m \leq l},     
\end{equation}
where $k^{l,m} \in \mathbb{R}^8$, 8 being the desired final feature dimension. We use 3 SH bands hence $l_{max} = 2$. 

Evaluating a view dependent point feature consists of linearly combining the Spherical Harmonic basis functions $\Phi_l^m : [0,2\pi]^2 \rightarrow \mathbb{R}$ evaluated at the viewing angle corresponding to a viewing direction $\Vec{d}$. The rasterized point feature at pixel $(u,v)$, as introduced in equation \ref{equ:pt_ft}, can thus be finally expressed as:
\begin{equation}
k_{u,v}^{\text{pt}} = w_{u,v} \times \sum_l \sum_m k_{p_{u,v}}^{l,m} \Phi_l^m(\Vec{d}).
\end{equation}
We recall that in concordance with definitions in the previous section (\ref{sec:FeatureRasterization}), $p_{u,v}$ is the point-rasterized vertex at location $(u,v)$. 

Similarly, the rasterized mesh feature at pixel $(u,v)$, as introduced in equation \ref{equ:ms_ft}, can be expressed as:
\begin{equation}
k_{u,v}^{\text{mesh}} = \sum_l \sum_m (\alpha k_{p_i}^{l,m} +  \beta k_{p_j}^{l,m} + \gamma k_{p_k}^{l,m}) \Phi_l^m(\Vec{d}).
\end{equation}
Here again, $(p_i,p_j,p_k)$ is the triangle rasterized at pixel $(u,v)$, and $\alpha$, $\beta$, $\gamma$ are the barycentric coordinates of the ray intersection with that triangle. Figure \ref{fig:ani_feat} illustrates the former equation. 

We note the the view direction $\Vec{d}$ for a pixel $(u,v)$ can be expressed as a function of the target camera parameters as follows: 
\begin{equation}
\Vec{d} = RK^{-1}\begin{pmatrix} u\\ v\\ 1\\ \end{pmatrix}+T.
\end{equation}

The rasterized view-dependent feature images ($k_{u,v} = \left[k_{u,v}^{\text{pt}},k_{u,v}^{\text{mesh}}\right]$)
are subsequently fed to a U-Net~\cite{ronneberger2015u}  based convolutional renderer to obtain the final novel rendered images. We will be detailing upon this neural renderer next.

\begin{figure}[t!]
\begin{center}
\includegraphics[width=0.7\linewidth]{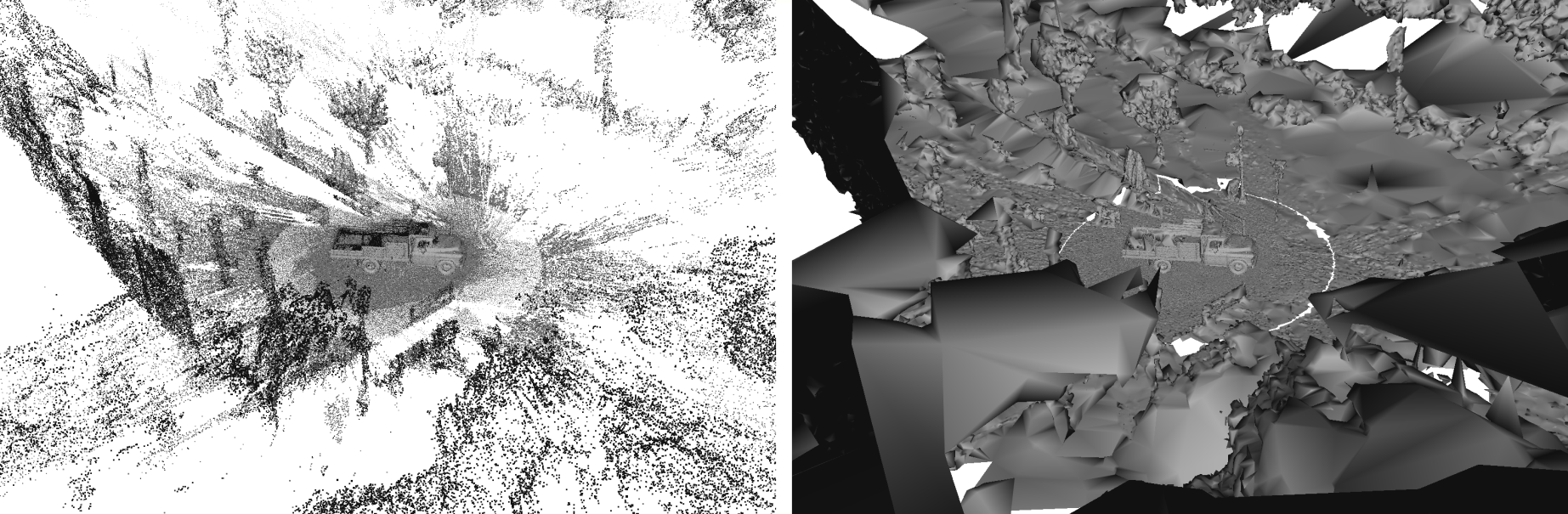}
\end{center}
   \caption{COLMAP~\cite{schonberger2016structure, schonberger2016pixelwise} geometry for scene ``Truck'' of Tanks and Temples~\cite{knapitsch2017tanks}. Left: Point cloud. Right: Mesh.}
\label{fig:rec}
\end{figure}

\begin{figure}[t!]
\begin{center}
\includegraphics[width=0.8\linewidth]{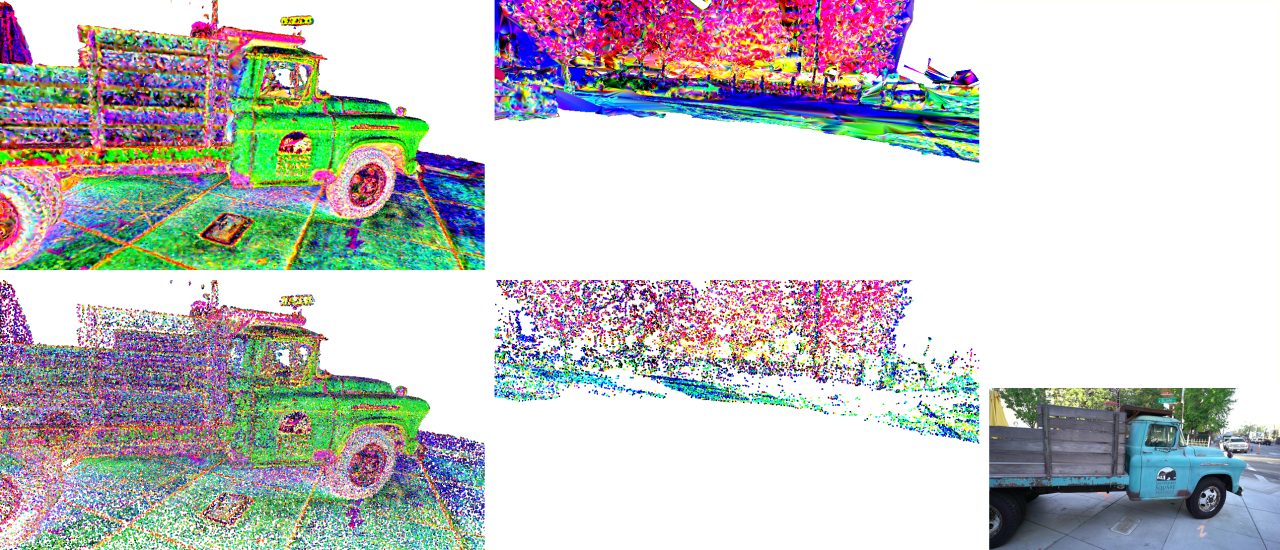}
\end{center}
   \caption{Left: Foreground rasterization. Right: background rasterization. Top row: Mesh based rasterization. Bottom row: Point based rasterization. We visualize feature PCA coefficients.}
\label{fig:split}
\end{figure}

\subsection{Split Neural Rendering}
\label{split_rendering}

Upon observing the geometry obtained from running COLMAP~\cite{schonberger2016structure, schonberger2016pixelwise}, especially on large unbounded scenes, such as the scenes in the Tanks and Temples dataset~\cite{knapitsch2017tanks}, the reconstruction is considerably more dense, detailed and precise for the main central area of interest where most cameras are pointing, as can be seen in Figure \ref{fig:rec}. The remaining of the scene geometry is sparse and less accurate. Hence, we argue that feature images rasterized from these foreground and background areas lie in two relatively separate domains, as the former is richer and more accurate than the latter. 

As such, we propose to split the proxy geometry into a foreground and background sub-meshes (\cf Figure \ref{fig:pipeline}).   The split is automatically performed following NeRF++~\cite{zhang2020nerf++}. Essentially, the center of the foreground sphere is approximated as the average camera position ($\bar{T}$), and its radius is set as 1.1 times the distance from this center to the furthest camera: $r_{fg} = \text{max}(||T-\bar{T}||_2)$. We separately rasterize the features from both areas (\cf Figure \ref{fig:split}), and we process these foreground and background feature images with two separate yet identical encoders, focusing each on processing feature images from their respective domains. While there are $2$ separate encoders, the features share a single decoder with skip connections coming in from both encoders. Other than these aspects, our neural renderer follows the multi-scale architecture in NPBG~\cite{aliev2020neural}.  




\subsection{Hybrid Training Objective}
\label{GAN_loss}

We experiment with a mix of supervised and self-supervised losses for training our method: $L=L_{\text{rec}} + L_{\text{GAN}}$. The loss is used to perform gradient descent on the parameters $\theta$ of the neural renderer $f_{\theta}$ and the scene point descriptors $\mathcal{K}$ jointly. 

\paragraph{Supervised loss}
The supervised loss follows the perceptual reconstruction loss in NPBG~\cite{aliev2020neural}, where we urge the network to reproduce the available groundtruth images ($I_{gt}$) in feature space, \ie:
\begin{equation}
L_{\text{rec}} = \sum_l||\Psi_{l}(f_{\theta}(\{k_{u,v}\})) - \Psi_{l}(I_{gt})||_{1}.
\end{equation}
$f_{\theta}(\{k_{u,v}\})$ is the output image from our network using rasterized features $\{k_{u,v}\}$. $\Psi_l$ represents the $l^{\text{th}}$ feature map from a pretrained VGG19 network~\cite{simonyan2014very}, where $l\in \{ \text{'conv1\_2','conv2\_2','conv3\_2', 'conv4\_2','conv5\_2'}\}$. 

\paragraph{Unsupervised loss}

We introduce a GAN~\cite{goodfellow2014generative} loss to encourage the photo-realism of the generated images and also improve generalization outside the training camera view-points. Specifically, with our entire model so far being the generator, we use a discriminator based on the DCGAN~\cite{radford2015unsupervised} model.
Besides sampling from the training cameras, we additionally sample artificial views following RegNeRF~\cite{niemeyer2022regnerf}. We note that for these augmented views we do not have a target ground-truth image, and hence only the GAN loss is back-propagated.  
To obtain the sample space of camera matrices, we assume that all cameras roughly focus on a central scene point $\bar{T}$, as described in Section~\ref{split_rendering}. The camera positions $\tilde{T}$ are sampled using a spherical coordinate system as follows: $\tilde{T} = \bar{T} + \tilde{r}[\sin{\theta}\sin{\phi},\cos{\theta},\sin{\theta}\cos{\phi}]$, where $\tilde{r}$ is sampled uniformly in
$[0.6 r_{fg}, r_{fg}]$, $r_{fg}$ being the foreground radius as defined in Section~\ref{split_rendering}. $\phi$ is sampled uniformly in $[0,2\pi]$. $\theta$ is sampled uniformly around the mean training camera elevation within $\pm$ 1.5 its standard deviation. For a given camera position, the camera rotation is defined using the camera ``look-at'' function, using target point $\bar{T}$ and the up axis of the scene.

\section{Results}\label{sec:Results}

\paragraph{Datasets}
To demonstrate the effectiveness of our approach for novel view synthesis in the context of large unbounded scenes, we choose the Tanks and Temples dataset~\cite{knapitsch2017tanks}. It consists of images in Full HD resolution captured by a high-end video camera. COLMAP~\cite{schonberger2016structure, schonberger2016pixelwise} is used to reconstruct the initial dense meshes/point clouds as well as to obtain the camera extrinsics and intrinsics for each scene. For quantitative evaluation, we follow the protocol in FVS~\cite{riegler2020free} and SVS~\cite{riegler2021stable}. $17$ of the $21$ scenes are used for training and for the rest of the scenes, i.e., ``Truck'', ``Train'', ``M60'', and ``Playground'', the images are divided into a fine-tuning set and a test set.
We also compare our method to prior approaches on the DTU~\cite{aanaes2016large} dataset, which consists of over $100$ tabletop scenes, with the camera poses being identical for all scenes. DTU~\cite{aanaes2016large} scenes are captured with a regular camera layout, which includes either $49$ or $64$ images with a resolution of $1200\times1600$ and their corresponding camera poses, taken from an octant of a sphere. We use scenes $65$, $106$ and $118$ for fine-tuning and testing purposes and the others are used for training. For scenes $65$, $106$ and $118$, of the total number of images and their corresponding camera poses, $10$ are selected as the testing set for novel view synthesis and the rest are used for fine-tuning. 

\paragraph{Training procedures}
We perform two types of training. In full training, we follow FVS~\cite{riegler2020free} and SVS~\cite{riegler2021stable} and jointly optimize the point neural descriptors of the training scenes along with the neural renderer. To test on a scene, we use the pretrained neural renderer and fine-tune it while learning the point descriptors of that scene, using the fine-tuning split. In single scene training, we perform the latter without pretraining the neural renderer.  

For single scene trainings, we train our network for $100$ epochs regardless of the scene. For full training, we pretrain our neural renderer for $15$ epochs on all training scenes followed by fine-tuning the entire network for $100$ epochs as before on specific scenes. For all our experiments, we use a batch size of $1$ with Adam~\cite{kingma2014adam} optimizer, and learning rates of $10^{-1}$ and $10^{-4}$ for the point neural descriptors and the neural renderer respectively.

\begin{figure*}[t!]
     \centering
     \begin{subfigure}[b]{0.2441\linewidth}
         \centering
         \includegraphics[width=\linewidth]{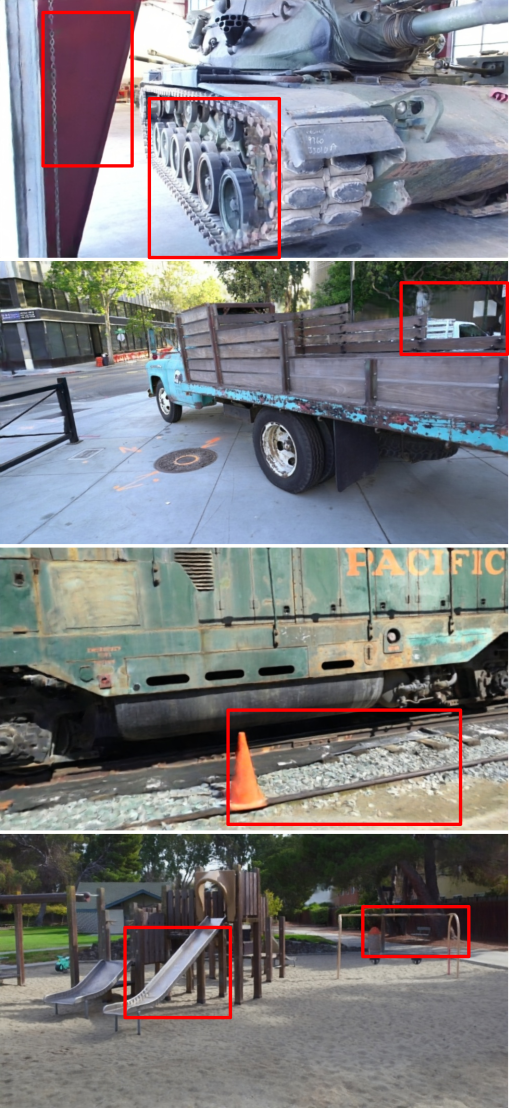}
        \caption{Ours}
     \end{subfigure}
     \hfill
     \begin{subfigure}[b]{0.2441\linewidth}
         \centering
         \includegraphics[width=\linewidth]{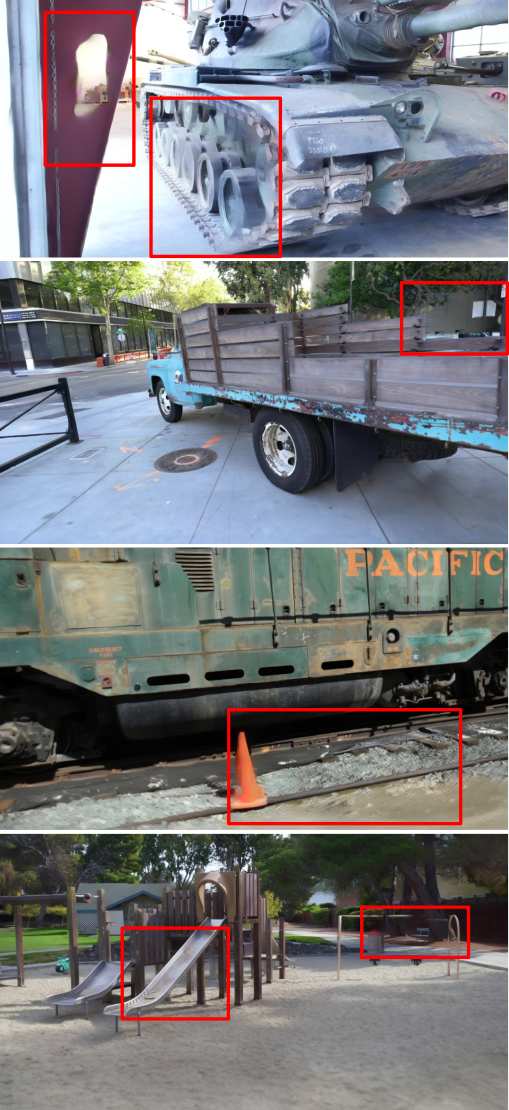}
        \caption{SVS~\cite{riegler2021stable}}
     \end{subfigure}
      \hfill
     \begin{subfigure}[b]{0.2441\linewidth}
         \centering
         \includegraphics[width=\linewidth]{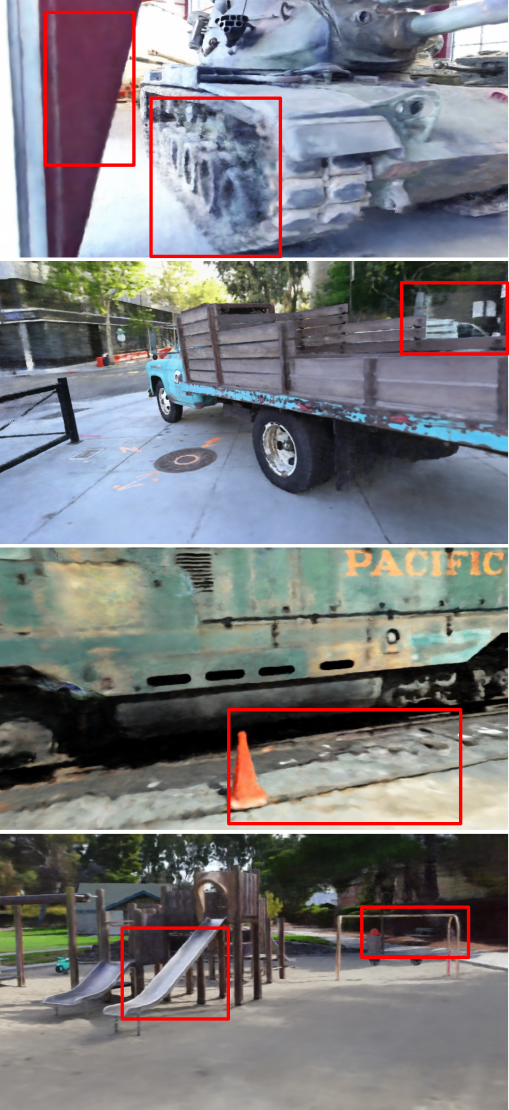}
        \caption{NeRF++~\cite{zhang2020nerf++}}
     \end{subfigure}
      \hfill
     \begin{subfigure}[b]{0.2441\linewidth}
         \centering
         \includegraphics[width=\linewidth]{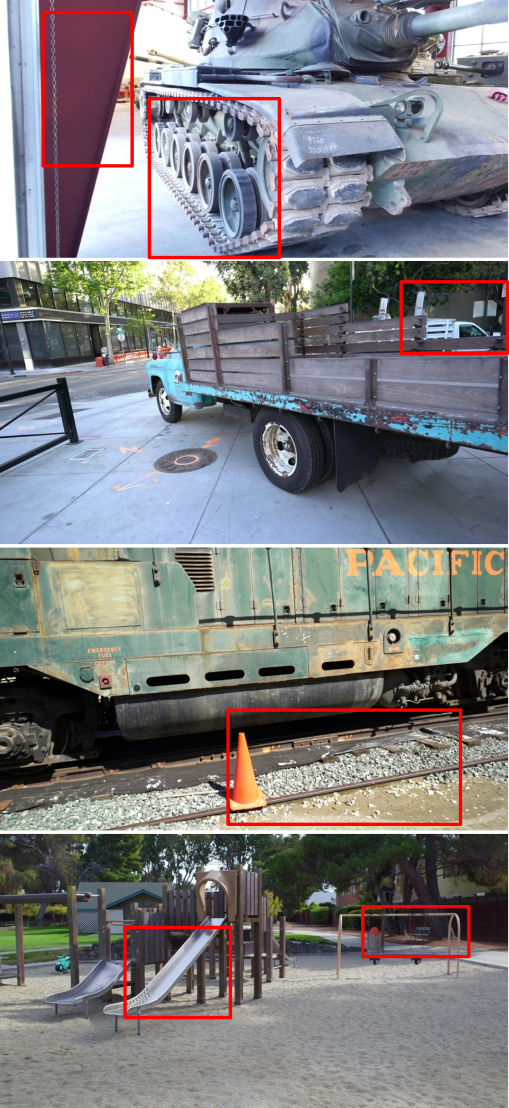}
        \caption{GT}
     \end{subfigure}
\caption{Qualitative comparison on Tanks and Temples \cite{knapitsch2017tanks}.}     
 \label{fig:qualitative_T_T}
\end{figure*}

\begin{table*}[h!]
  \centering
\scalebox{0.7}{
\begin{tabular}{ c|ccc|ccc|ccc|ccc}
 \hline
 Methods&\multicolumn{3}{c|}{Truck}& \multicolumn{3}{c|}{M60}& \multicolumn{3}{c|}{Playground}& \multicolumn{3}{c}{Train}\\ 
 & PSNR $\uparrow$ &SSIM $\uparrow$ &LPIPS $\downarrow$ & PSNR $\uparrow$ &SSIM $\uparrow$ &LPIPS $\downarrow$ & PSNR $\uparrow$ &SSIM $\uparrow$ &LPIPS $\downarrow$ & PSNR $\uparrow$ &SSIM $\uparrow$ &LPIPS $\downarrow$\\
 \hline
 LLFF~\cite{mildenhall2019local} &10.78 &0.454 &60.62 &8.98 &0.431 &71.76 &14.40 &0.578 &53.93 &9.15 &0.384 &67.40\\
 EVS~\cite{choi2019extreme} &14.22 &0.527 &43.52 &7.41 &0.354 &75.71 &14.72 &0.568 &46.85 &10.54 &0.378 &67.62\\
 NeRF~\cite{mildenhall2020nerf} &20.85 &0.738 &50.74 &16.86 &0.701 &60.89 &21.55 &0.759 &52.19 &16.64 &0.627 &64.64\\
 NeRF++~\cite{zhang2020nerf++} &22.77 &0.814 &30.04 &18.49 &0.747 &43.06 &22.93 &0.806 &38.70 &17.77 &0.681 &47.75\\
 FVS~\cite{riegler2020free} &22.93 &0.873 &13.06 &16.83 &0.783 &30.70 &22.28 &0.846 &19.47 &18.09 &0.773 &24.74\\
 SVS~\cite{riegler2021stable} &23.86 &0.895 &9.34 &19.97 &0.833 &20.45 &23.72 &0.884 &14.22 &18.69 &0.820 &15.73\\
 \hline
 NPBG~\cite{aliev2020neural} &21.88 &0.877 &\cellcolor{shade4}15.04 &12.35 &0.716 &35.57 &23.03 &\cellcolor{shade4}0.876 &\cellcolor{shade2}16.65 &\cellcolor{shade4}18.08 &\cellcolor{shade4}0.801 &25.48\\
 Ours (Single) &\cellcolor{shade5}23.88 &\cellcolor{shade5}0.883 &17.41 &\cellcolor{shade4}19.34 &\cellcolor{shade4}0.810 &\cellcolor{shade4}24.13 &\cellcolor{shade5}23.38 &0.865 &23.34 &17.35 &0.788 &\cellcolor{shade4}23.66\\
 Ours (Full) &\cellcolor{shade4}24.03 &\cellcolor{shade4}0.888 &\cellcolor{shade5}16.84 &\cellcolor{shade2}19.54 &\cellcolor{shade2}0.815 &\cellcolor{shade2}23.15 &\cellcolor{shade4}23.59 &\cellcolor{shade5}0.870 &\cellcolor{shade5}22.72 &\cellcolor{shade5}17.78 &\cellcolor{shade5}0.799 &\cellcolor{shade5}24.17\\
 \hline
\end{tabular}
}
\vspace{10pt}
\caption{Quantitative comparison on Tanks and Temples~\cite{knapitsch2017tanks}. Deeper shades represent better performance.} \label{tab:table1}
\vspace{-25pt}
\end{table*}

\paragraph{Metrics} We report our performance for view synthesis, in line with previous seminal work, using three image fidelity metrics, namely Peak signal-to-noise ratio (PSNR), structural similarity (SSIM)~\cite{wang2004image} and learned perceptual image patch similarity (LPIPS)~\cite{zhang2018unreasonable}.

\paragraph{Quantitative comparison}

For our method, we show both single scene training (Ours (Single)) and full dataset training (Our (Full)) results. We relay the performances of methods \cite{mildenhall2019local, choi2019extreme, aliev2020neural, mildenhall2020nerf, zhang2020nerf++, riegler2020free, riegler2021stable} as reported in SVS~\cite{riegler2021stable}.

Table \ref{tab:table1} shows a quantitative comparison of our method with the recent state-of-the-art on Tanks and Temples~\cite{knapitsch2017tanks}. Most methods underperform on these challenging large unbounded scenes apart from NPBG~\cite{aliev2020neural}, FVS~\cite{riegler2020free} and SVS~\cite{riegler2021stable}.
Although they could achieve promising results with only single scene training, methods based on volumetric neural rendering (NeRF~\cite{mildenhall2020nerf} and NeRF++~\cite{zhang2020nerf++}) are famously computationally expensive to train and render.  
Even by training on a single scene only, our method outperforms NPBG~\cite{aliev2020neural} in almost all scenes. We note that the neural renderer of NPBG~\cite{aliev2020neural} here was pretrained on ScanNet dataset \cite{dai2017scannet}.  
Our method produces competitive results with respect to the best performing methods on this benchmark, \ie FVS~\cite{riegler2020free} and SVS~\cite{riegler2021stable}. It is interesting to observe in particular that with merely single scene training, our method outperforms the state-of-the-art SVS on scene ``Truck'', while coming as a close second in almost all other scenes in PSNR and SSIM~\cite{wang2004image}. 
FVS~\cite{riegler2020free} and SVS~\cite{riegler2021stable} perform exceedingly well in all metrics but require training on the entire dataset, with a considerably larger training time than our single scene training. In particular, their lower LPIPS values could be attributed to the use of a considerably deeper neural renderer than ours, consisting of $9$ consecutive U-Nets~\cite{ronneberger2015u}. Ours is a much lighter single U-Net~\cite{ronneberger2015u}. 

\begin{figure*}[t!]
     \centering
     \begin{subfigure}[t]{0.22\linewidth}
         \centering
         \includegraphics[width=\linewidth]{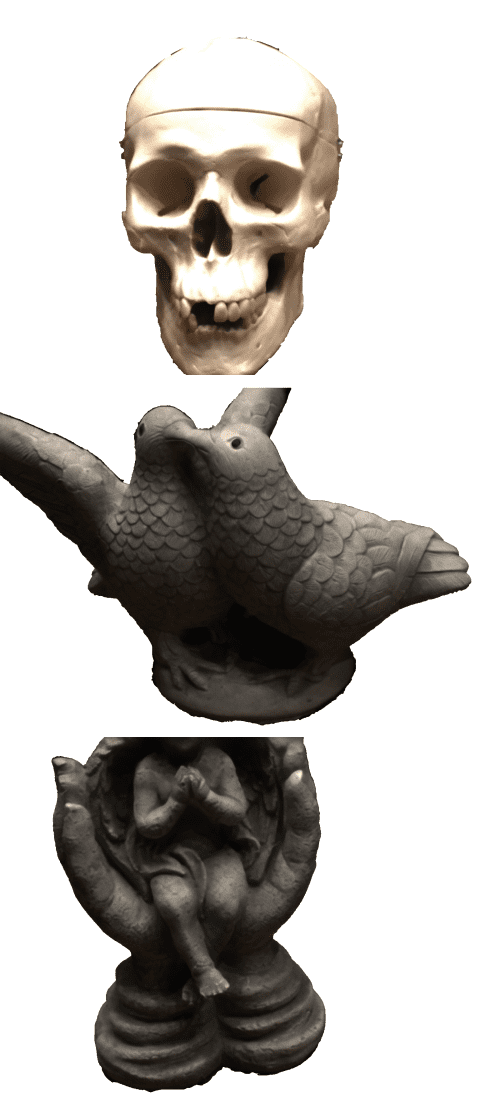}
         \captionsetup{font=tiny,labelfont=tiny}
        \caption{Ours (Full)}
     \end{subfigure}
     \begin{subfigure}[t]{0.22\linewidth}
         \centering
         \includegraphics[width=\linewidth]{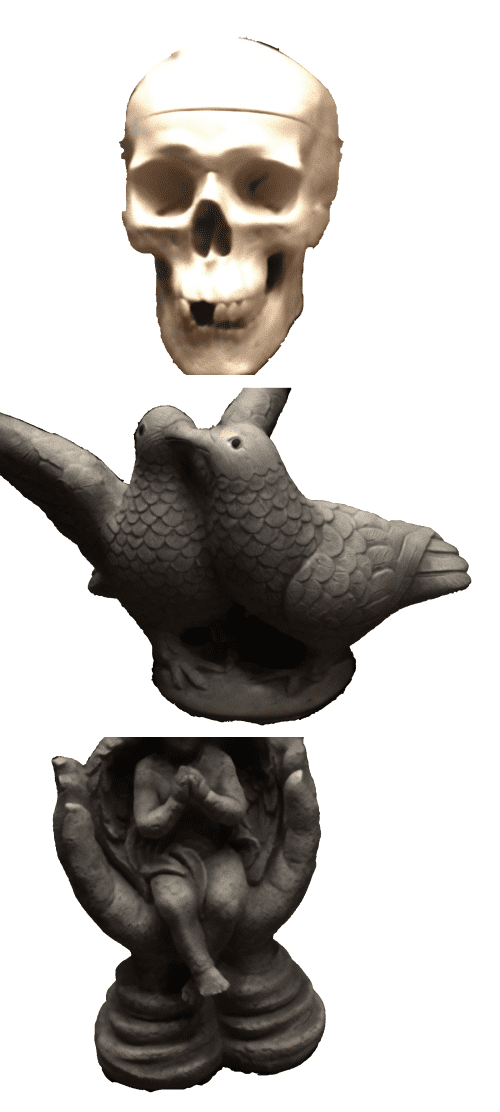}
         \captionsetup{font=tiny,labelfont=tiny}
        \caption{Ours (Single)}
     \end{subfigure}
     \begin{subfigure}[t]{0.22\linewidth}
         \centering
         \includegraphics[width=\linewidth]{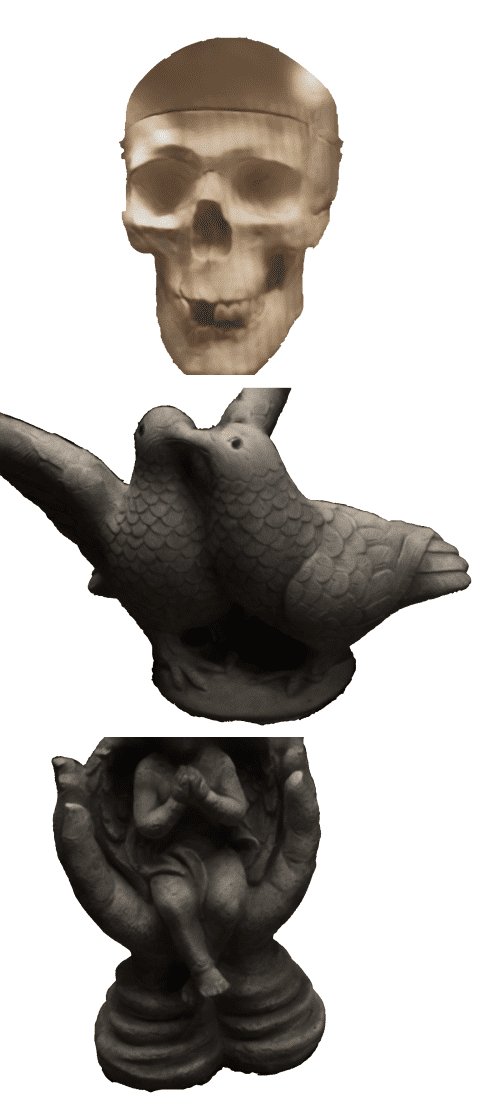}
         \captionsetup{font=tiny,labelfont=tiny}
        \caption{NPBG~\cite{aliev2020neural} (Full)}
     \end{subfigure}
     \begin{subfigure}[t]{0.22\linewidth}
         \centering
         \includegraphics[width=\linewidth]{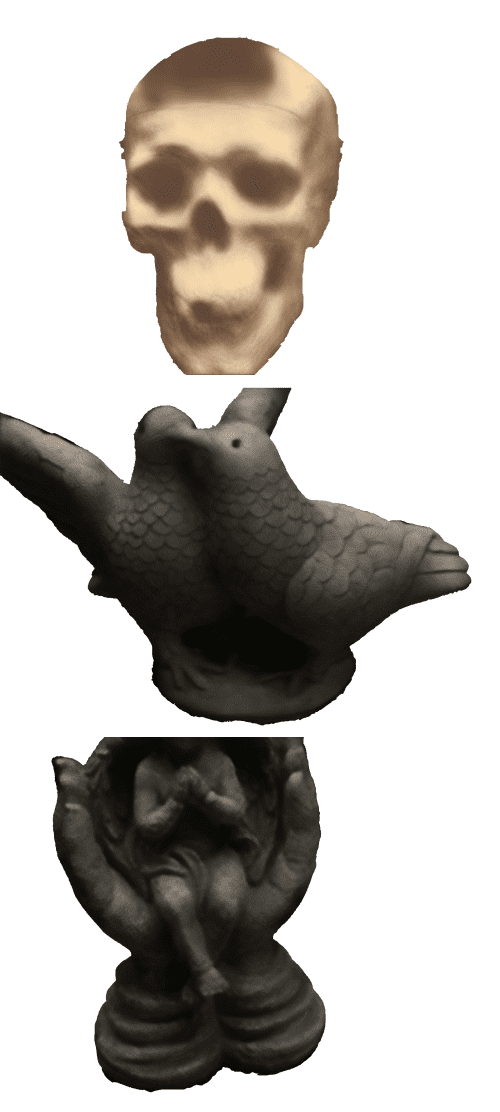}
         \captionsetup{font=tiny,labelfont=tiny}
        \caption{NPBG~\cite{aliev2020neural} (Single)}
     \end{subfigure}
 \caption{Qualitative comparison on DTU~\cite{aanaes2016large}.}     
 \label{fig:qualitative_DTU}
\end{figure*}
\begin{table*}[t!]
  \centering
\scalebox{0.68}{
\begin{tabular}{ c|ccc|ccc|ccc}
 \hline
 Methods&\multicolumn{3}{c|}{65}& \multicolumn{3}{c|}{106}& \multicolumn{3}{c}{118}\\ 
 & PSNR $\uparrow$ &SSIM $\uparrow$ &LPIPS $\downarrow$ & PSNR $\uparrow$ &SSIM $\uparrow$ &LPIPS $\downarrow$ & PSNR $\uparrow$ &SSIM $\uparrow$ &LPIPS $\downarrow$\\
 \hline
 LLFF~\cite{mildenhall2019local} &22.48/22.07 &0.935/0.921 &9.38/12.71 &24.10/24.63 &0.900/0.886 &13.26/13.57 & 28.99/27.42 &0.928/0.922 &9.69/10.99\\
 EVS~\cite{choi2019extreme} &23.26/14.43 &0.942/0.848 &7.94/22.11 &20.21/11.15 &0.902/0.743 &14.91/29.57 &23.35/12.06 &0.928/0.793 &10.84/25.01\\
 NeRF~\cite{mildenhall2020nerf} &32.00/28.12 &0.984/0.963 &3.04/8.54 &34.45/30.66 &0.975/0.957 &7.02/10.14 &37.36/31.66 &0.985/0.967 &4.18/6.92\\
 FVS~\cite{riegler2020free} &30.44/25.32 &0.984/0.961 &2.56/7.17 &32.96/27.56 &0.979/0.950 &2.96/6.57 &35.64/29.54 &0.985/0.963 &1.95/6.31\\
 SVS~\cite{riegler2021stable} &32.13/26.82 &0.986/0.964 &1.70/5.61 &34.30/30.64 &0.983/0.965 &1.93/3.69 &37.27/31.44 &0.988/0.967 &1.30/4.26\\
 \hline
 NPBG~\cite{aliev2020neural} &16.74/15.44 &0.889/0.873 &14.30/19.45 &19.62/20.26 &0.847/0.842 &18.90/21.13 &23.81/24.14 &0.867/0.879 &15.22/16.88\\
 Ours (Single) &\cellcolor{shade4}26.78/20.85 &\cellcolor{shade4}0.957/0.925 &\cellcolor{shade4}9.64/12.91 &\cellcolor{shade4}29.98/25.40 &\cellcolor{shade4}0.931/0.909 &\cellcolor{shade4}12.75/13.70 &\cellcolor{shade4}31.43/26.52 &\cellcolor{shade4}0.946/0.931 &\cellcolor{shade4}11.73/11.13\\
 Ours (Full) &\cellcolor{shade2}28.98/22.90 &\cellcolor{shade2}0.970/0.943 &\cellcolor{shade2}7.15/11.13 &\cellcolor{shade2}30.67/25.75 &\cellcolor{shade2}0.939/0.917 &\cellcolor{shade2}12.10/13.10 &\cellcolor{shade2}32.39/27.97 &\cellcolor{shade2}0.956/0.941 &\cellcolor{shade2}10.62/10.07\\
 \hline
\end{tabular}
}
\captionsetup{font=small,labelfont=small}
\vspace{10pt}
\caption{Quantitative comparison on DTU~\cite{aanaes2016large}. Left/Right: View interpolation/extrapolation. Deeper shades represent better performance.} \label{tab:table4}
\vspace{-25pt}
\end{table*}

Table \ref{tab:table4} reports quantitative comparison on the DTU~\cite{aanaes2016large} dataset. We use the view interpolation and extrapolation setting, as adopted by SVS~\cite{riegler2021stable} and FVS~\cite{riegler2020free} for a fair comparison with other methods. The setting comprises of $6$ central cameras to evaluate view interpolation and $4$ corner cameras to evaluate view extrapolation. As has been observed from the experiments conducted in SVS~\cite{riegler2021stable}, LLFF~\cite{mildenhall2019local} and EVS~\cite{choi2019extreme} perform decently on DTU~\cite{aanaes2016large}, while NeRF~\cite{mildenhall2020nerf}, FVS~\cite{riegler2020free} and SVS~\cite{riegler2021stable} excel on it. NPBG~\cite{aliev2020neural}, on the other hand, performs very poorly due to lack of data per scene (approximately 39 images) making the point feature autodecoding less efficient. Our method, despite using point feature autodecoding equally, gains considerably on NPBG~\cite{aliev2020neural} and performs relatively close to FVS~\cite{riegler2020free}. Again, as observed for the Tanks and Temples~\cite{knapitsch2017tanks} dataset, our method is able to rapidly train for a single scene and perform relatively well, and close to methods which have been trained and finetuned on the entire dataset such as FVS~\cite{riegler2020free} and SVS~\cite{riegler2021stable}, which makes it very practical in the sense that it is able to achieve reasonable results with limited training time and data.


\begin{figure}[t!]
     \centering
     \begin{subfigure}[b]{0.200\linewidth}
         \centering
         \includegraphics[width=\linewidth]{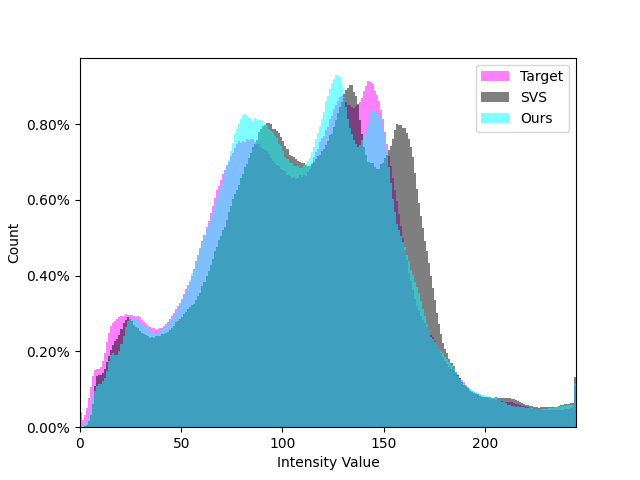}
     \end{subfigure}
     \begin{subfigure}[b]{0.200\linewidth}
         \centering
         \includegraphics[width=\linewidth]{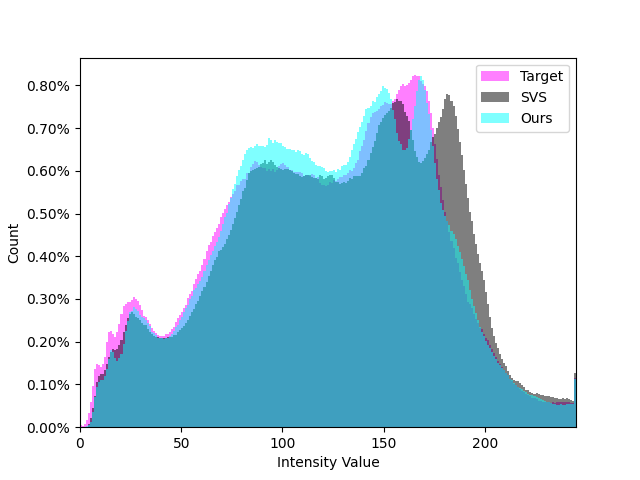}
     \end{subfigure}
     \begin{subfigure}[b]{0.200\linewidth}
         \centering
         \includegraphics[width=\linewidth]{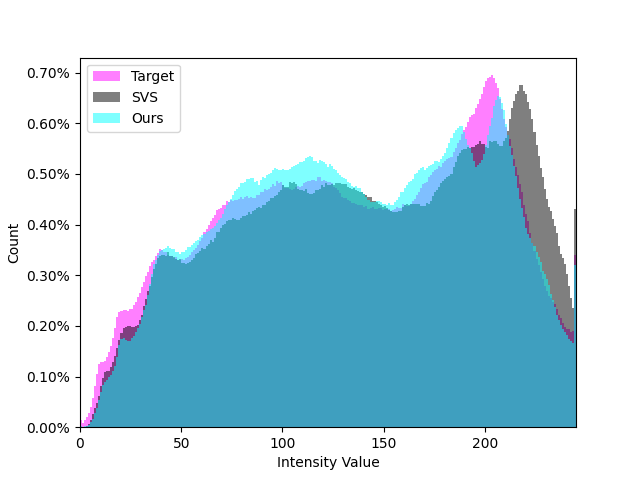}
     \end{subfigure}
 \caption{Red, Blue and Green histograms for images in the ``Truck'' scene of Tanks and Temples~\cite{knapitsch2017tanks} for our method,  SVS~\cite{riegler2021stable} and the groundtruth.}
 \label{fig:color_hist}
 \vspace{-10pt}
\end{figure}

\paragraph{Qualitative comparison}

Qualitative results on Tanks and Temples~\cite{knapitsch2017tanks} of some of the competitive contemporary methods are summarized in Figure \ref{fig:qualitative_T_T}. In general, we notice that SVS~\cite{riegler2021stable} excels in the synthesis quality of the novel views. This is also reflected in the LPIPS~\cite{zhang2018unreasonable} metric, where SVS~\cite{riegler2021stable} performs better than the competition. NeRF++~\cite{zhang2020nerf++} on the other hand tends to underperform and produces blurs and artifacts, particularly if we look at the results of the ``M60'' and ``Train'' scenes. 

Although SVS~\cite{riegler2021stable} performs well overall, especially in background regions, it tends to display an  unnatural ``smoothing effect'' on the image in certain regions, such as the rocks indicated in the ``Train'' scene or in the track of the tank in the ``M60'' scene. We suspect that this is due to their neural renderer which contains $9$ consecutive U-Nets~\cite{ronneberger2015u}, which might cause the view-dependent feature tensor to be oversmoothed. Furthermore, SVS~\cite{riegler2021stable} results sometimes contain major artifacts like holes, missing structures or transparent parts in certain regions, such as the ones indicated in the ``M60'', ``Truck'' and ``Playground'' scenes respectively in Figure \ref{fig:qualitative_T_T}. Another potential issue that we observed was an apparent color shift in some of the results of SVS~\cite{riegler2021stable}. To investigate this, we plot separate color histograms of the synthesized outputs of SVS~\cite{riegler2021stable}, our method and the target images for the ``Truck'' scene in Figure \ref{fig:color_hist}. We notice in this figure that the histogram of the images synthesized by our method matches that of the target images more closely than those synthesized by SVS~\cite{riegler2021stable}. Overall, our method tends to display less of the color shifts and holes/artifacts that SVS~\cite{riegler2021stable} seems to exhibit despite lower scores with respect to the evaluation metrics, particularly for LPIPS~\cite{zhang2018unreasonable}.

Next, we present some qualitative results on DTU~\cite{aanaes2016large} in Figure \ref{fig:qualitative_DTU}. We notice that compared to NPBG~\cite{aliev2020neural} which generates very poor quality novel views (both for single-scene finetuning and full training) presumably due to lack of adequate training data, our method offers way better visual results which we believe is due to a denser input feature image fed to the neural renderer, on account of point based rasterization with bigger radius and the mesh based rasterization. In fact, our method, with just single-scene finetuning performs better than NPBG~\cite{aliev2020neural} even when fully trained, which is also supported by the quantitative results presented in Table \ref{tab:table4}.

\paragraph{Ablation studies}

\begin{table*}[t!]
 \centering
\scalebox{0.7}{
\begin{tabular}{ c|ccc|ccc|ccc|ccc}
 \hline
 Methods&\multicolumn{3}{c|}{Truck}& \multicolumn{3}{c|}{M60}& \multicolumn{3}{c|}{Playground}& \multicolumn{3}{c}{Train}\\ 
 & PSNR $\uparrow$ &SSIM $\uparrow$ &LPIPS $\downarrow$ & PSNR $\uparrow$ &SSIM $\uparrow$ &LPIPS $\downarrow$ & PSNR $\uparrow$ &SSIM $\uparrow$ &LPIPS $\downarrow$ & PSNR $\uparrow$ &SSIM $\uparrow$ &LPIPS $\downarrow$\\
 \hline
 Original NPBG &21.55 &0.807 &27.58 &17.48 &0.757 &33.39 &22.26 &0.808 &28.79 &16.08 &0.697 &34.58\\
 Bigger radius &22.26 &0.832 &23.55 &18.76 &0.784 &28.73 &22.61 &0.821 &26.62 &16.15 &0.717 &30.89\\
 With mesh &22.70 &0.850 &20.76 &{\bf 19.95} &\underline{0.811} &24.26 &22.78 &0.838 &25.22 &15.75 &0.730 &29.01\\
 Dir. features &23.40 &0.872 &18.57 &19.26 &0.810 &\underline{24.05} &23.27 &0.857 &24.42 &17.13 &0.782 &24.79\\
 Split scene &\underline{23.64} &\underline{0.880} &\underline{17.99} &19.03 &{\bf 0.812} &{\bf 24.04} &\underline{23.32} &\underline{0.865} &{\bf 23.11} &{\bf 17.37} &\underline{0.787} &\underline{24.13}\\
 Ours (Single) &{\bf 23.88} &{\bf 0.883} &{\bf 17.41} &\underline{19.34} &0.810 &24.13 &{\bf 23.38} &{\bf 0.865} &\underline{23.34} &\underline{17.35} &{\bf 0.788} &{\bf 23.66}\\
 \hline
\end{tabular}
}
\vspace{10pt}
\captionsetup{font=small,labelfont=small}
\caption{Quantitative ablation on Tanks and Temples~\cite{knapitsch2017tanks}. Best/second best performances are emboldened/underlined respectively.} \label{tab:table3}
\vspace{-25pt}
\end{table*}

In this section, we conduct an ablative analysis to justify the choice of our final architecture. We ablate in the single-scene training scenario using all testing scenes of Tanks and Temples~\cite{knapitsch2017tanks}. 
We progressively add components to our baseline architecture (i.e. NPBG~\cite{aliev2020neural}) until we reach our final model to demonstrate their individual contributions to our performance. The results are summarized in Table \ref{tab:table3}. ``Original NPBG'' is NPBG~\cite{aliev2020neural} in the single-scene setting. ``Bigger radius'' is the ``Original NPBG'' with a bigger rasterization radius as discussed in Section \ref{sec:FeatureRasterization}, as opposed to NPBG~\cite{aliev2020neural} which has a rasterization radius of half a pixel. ``With mesh'' includes the mesh rasterized and interpolated feature image mentioned in Section \ref{sec:FeatureRasterization}. These two previous components lead to denser feature images.
``Directional features'' incorporates view dependency in the geometry descriptors using Spherical Harmonics (SH) as discussed in Section \ref{sec:DirnFeatures}, while ``Split scene'' splits the proxy geometry into foreground and background, rasterizes and encodes each features separately. Our final model uses an additional GAN~\cite{goodfellow2014generative} loss during training as described in Section \ref{GAN_loss}. Overall throughout all scenes, the numbers witness the consistent improvement brought by the various components. 

\section{Conclusion}

We improved in this work on the Neural Point Based Graphics (i.e. NPBG~\cite{aliev2020neural}) model for novel view synthesis, by providing a new data-efficient version that can achieve superior results by training solely on a single scene. SVS~\cite{riegler2021stable} still produces Superior LPIPS~\cite{zhang2018unreasonable} synthesis performance. As future work, we will investigate and improve on this aspect of our method, while attempting to maintain memory and compute efficiency. 

\clearpage
%
%
\bibliographystyle{splncs04}
\bibliography{egbib}
\end{document}


\pagestyle{headings}
\mainmatter
\def\ECCVSubNumber{24}  
\title{Neural Mesh-Based Graphics\\– Supplementary Material –}

\author{\large Shubhendu Jena, Franck Multon, Adnane Boukhayma}
\authorrunning{S. Jena et al.}
%
\institute{\normalsize Inria, Univ. Rennes, CNRS, IRISA, M2S, France}
%

%
%
\maketitle

\section{Additional qualitative results}
\begin{figure*}[h!]
     \centering
     \begin{subfigure}[b]{0.2441\linewidth}
         \centering
         \includegraphics[width=\linewidth]{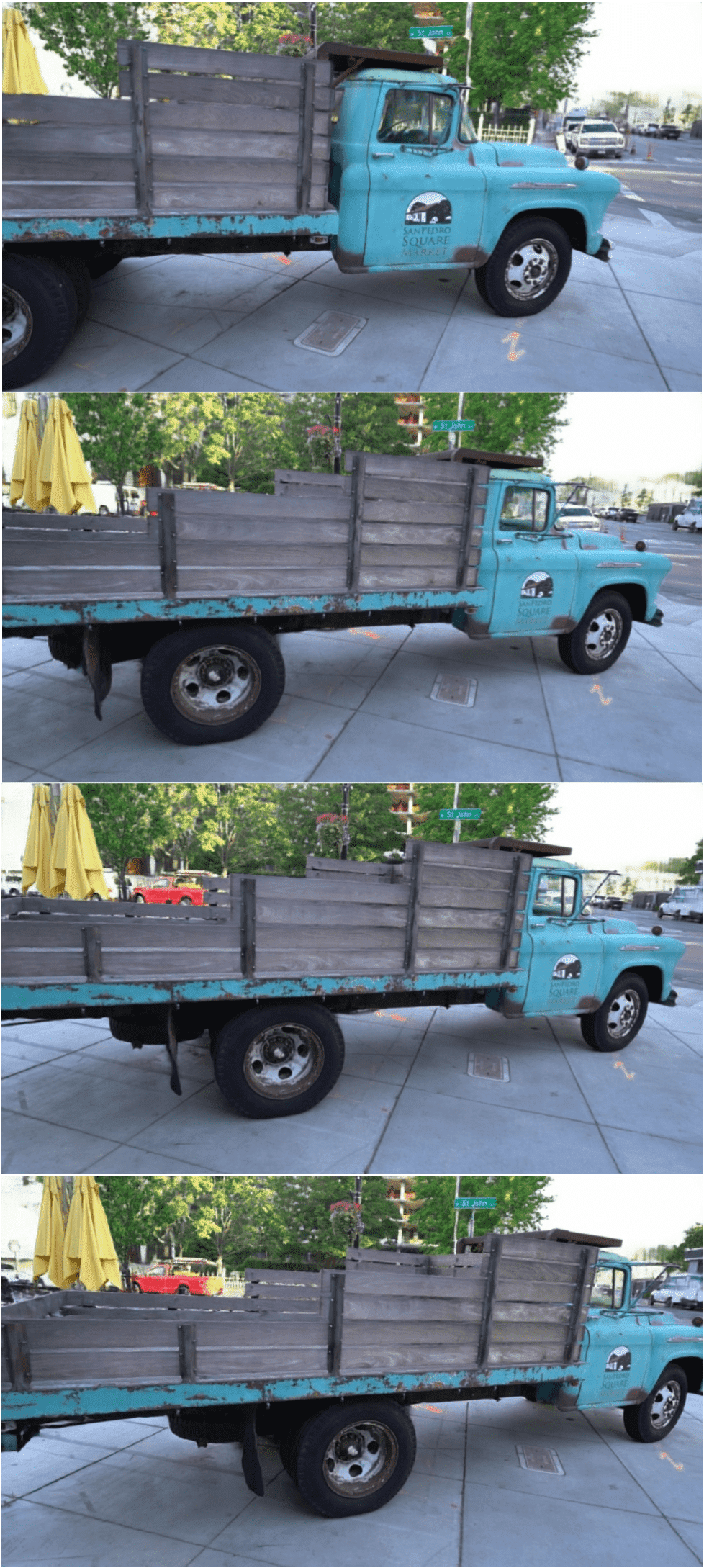}
     \end{subfigure}
     \hfill
     \begin{subfigure}[b]{0.2441\linewidth}
         \centering
         \includegraphics[width=\linewidth]{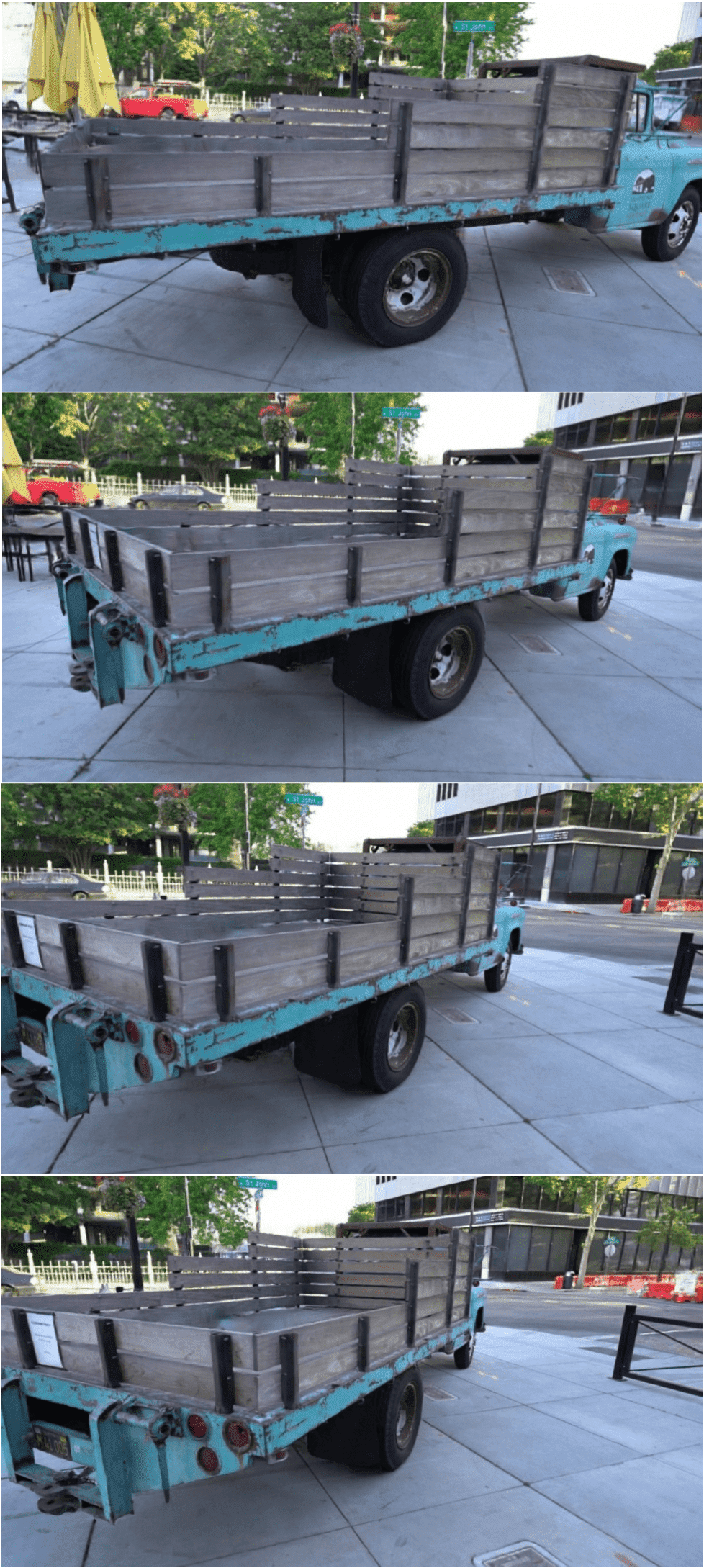}
     \end{subfigure}
      \hfill
     \begin{subfigure}[b]{0.2441\linewidth}
         \centering
         \includegraphics[width=\linewidth]{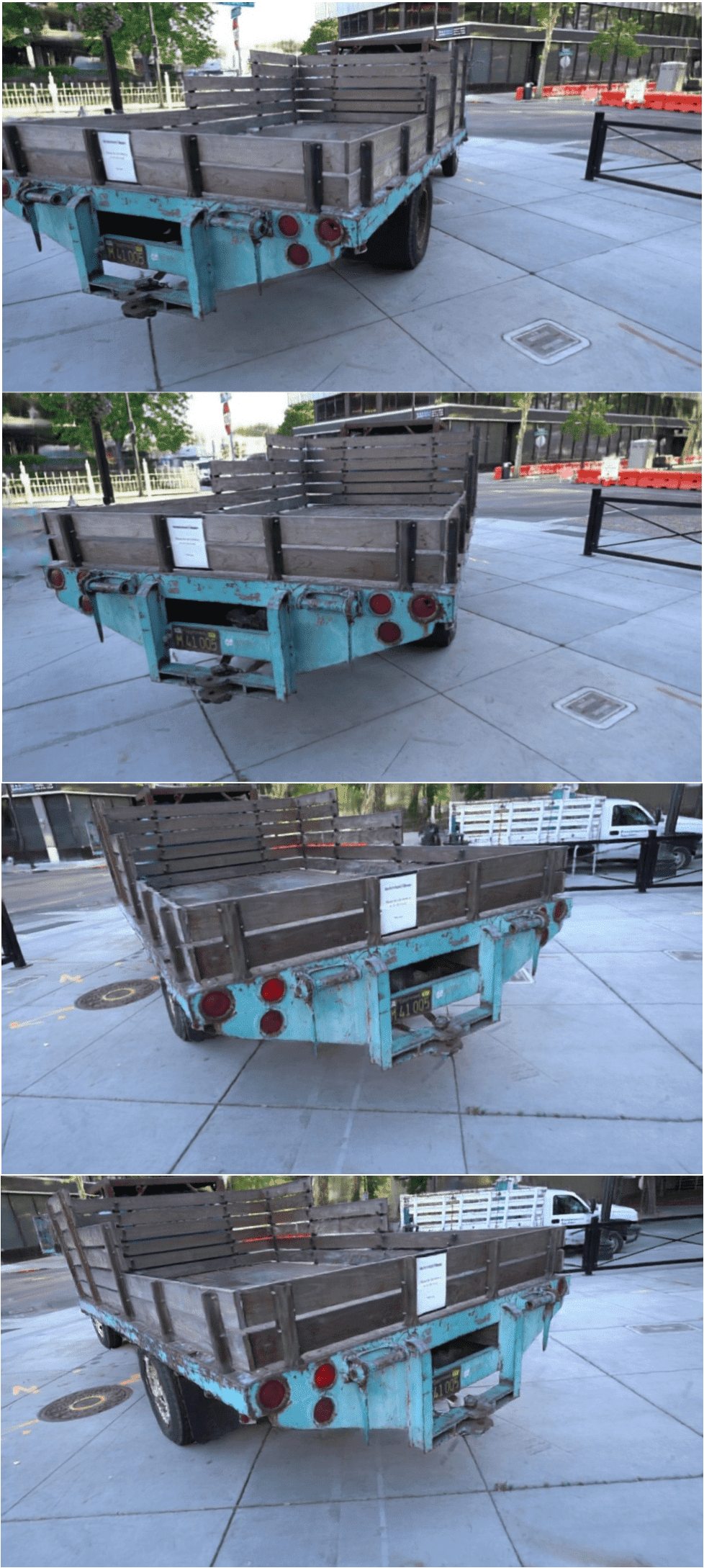}
     \end{subfigure}
      \hfill
     \begin{subfigure}[b]{0.2441\linewidth}
         \centering
         \includegraphics[width=\linewidth]{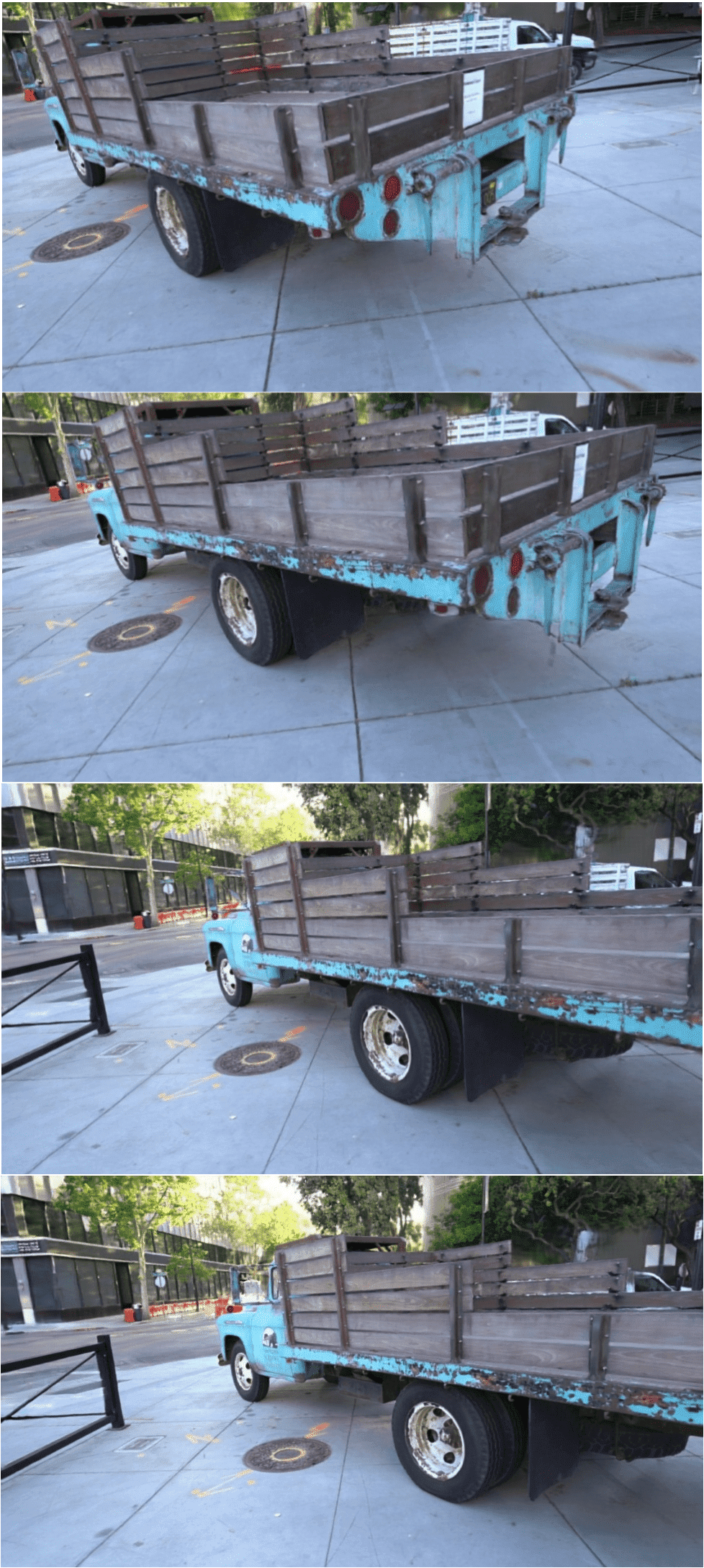}
     \end{subfigure}
\caption{Additional results on ``Truck'' scene of the Tanks and Temples \cite{knapitsch2017tanks} dataset.}     
 \label{fig:qualitative_T_T_1}
\end{figure*}

\begin{figure*}[h!]
     \centering
     \begin{subfigure}[b]{0.2441\linewidth}
         \centering
         \includegraphics[width=\linewidth]{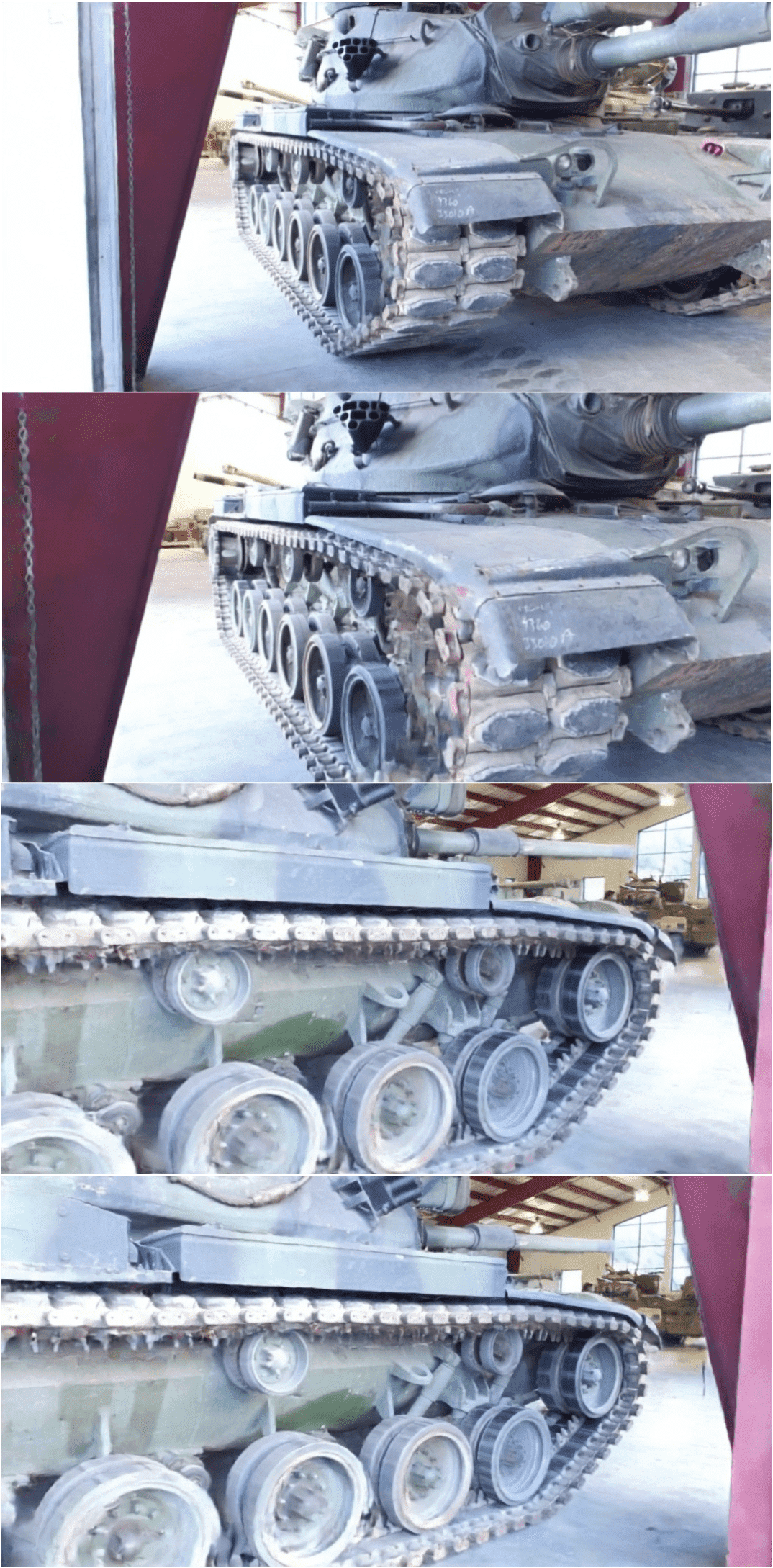}
     \end{subfigure}
     \hfill
     \begin{subfigure}[b]{0.2441\linewidth}
         \centering
         \includegraphics[width=\linewidth]{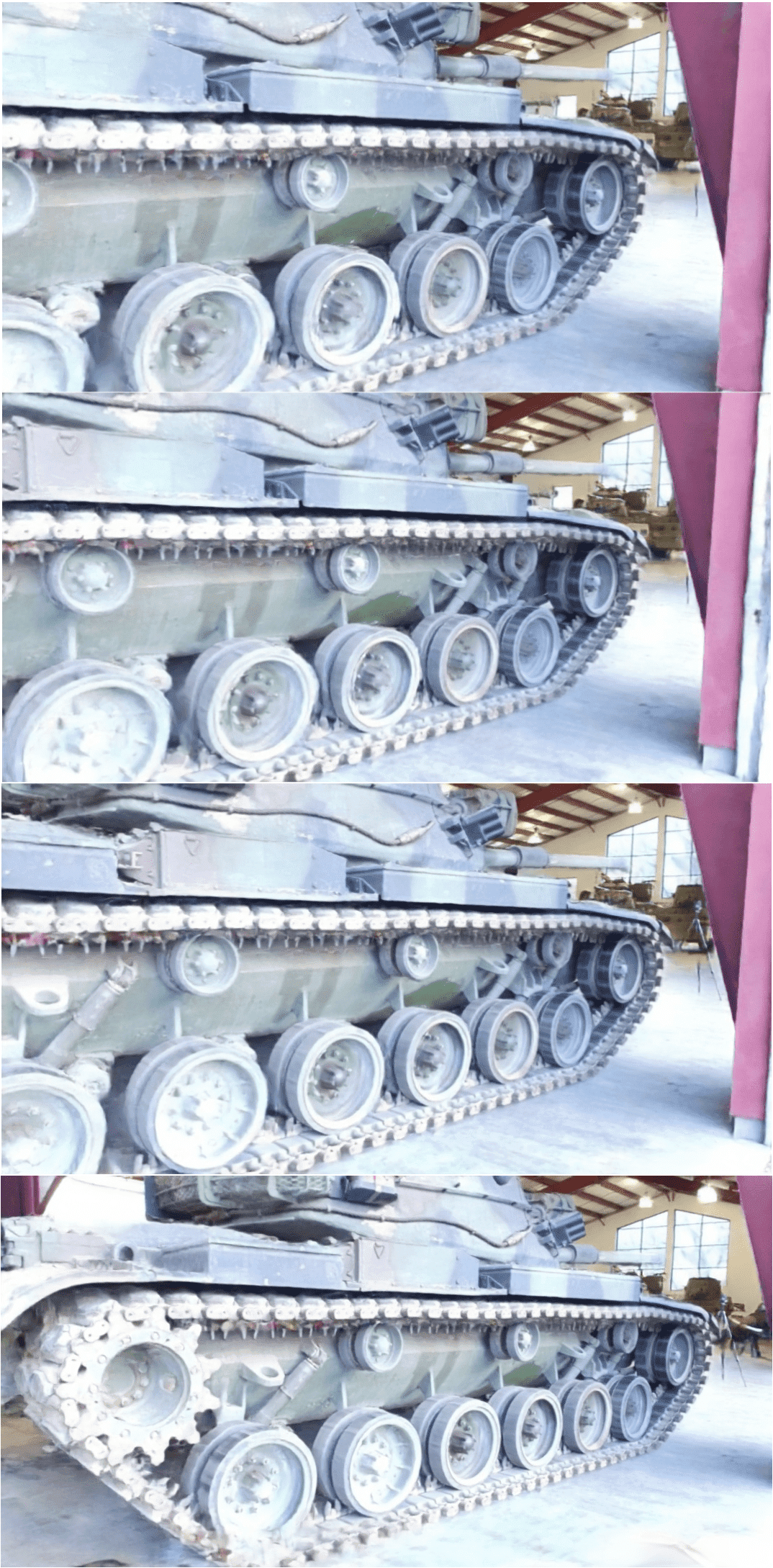}
     \end{subfigure}
      \hfill
     \begin{subfigure}[b]{0.2441\linewidth}
         \centering
         \includegraphics[width=\linewidth]{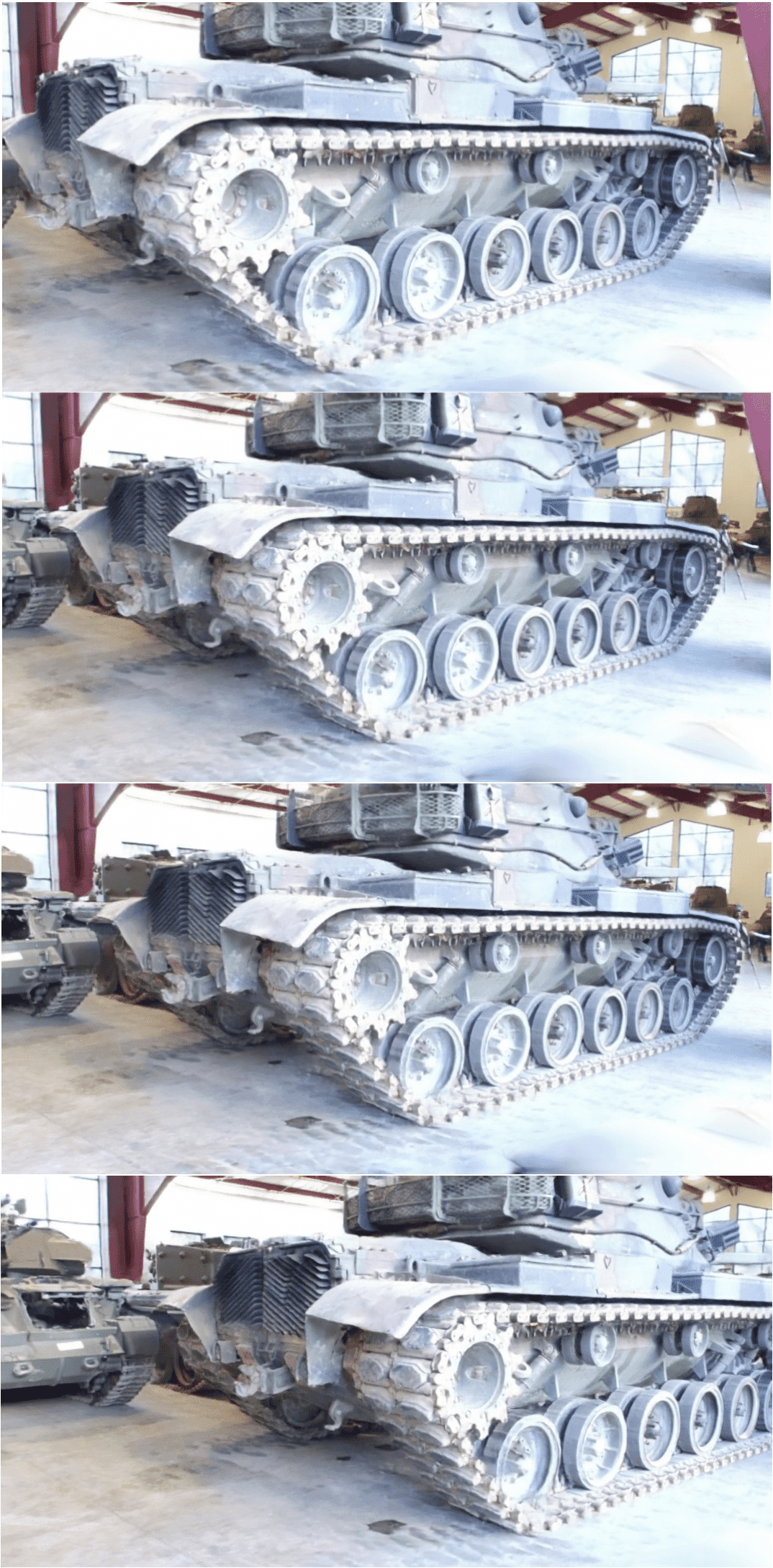}
     \end{subfigure}
      \hfill
     \begin{subfigure}[b]{0.2441\linewidth}
         \centering
         \includegraphics[width=\linewidth]{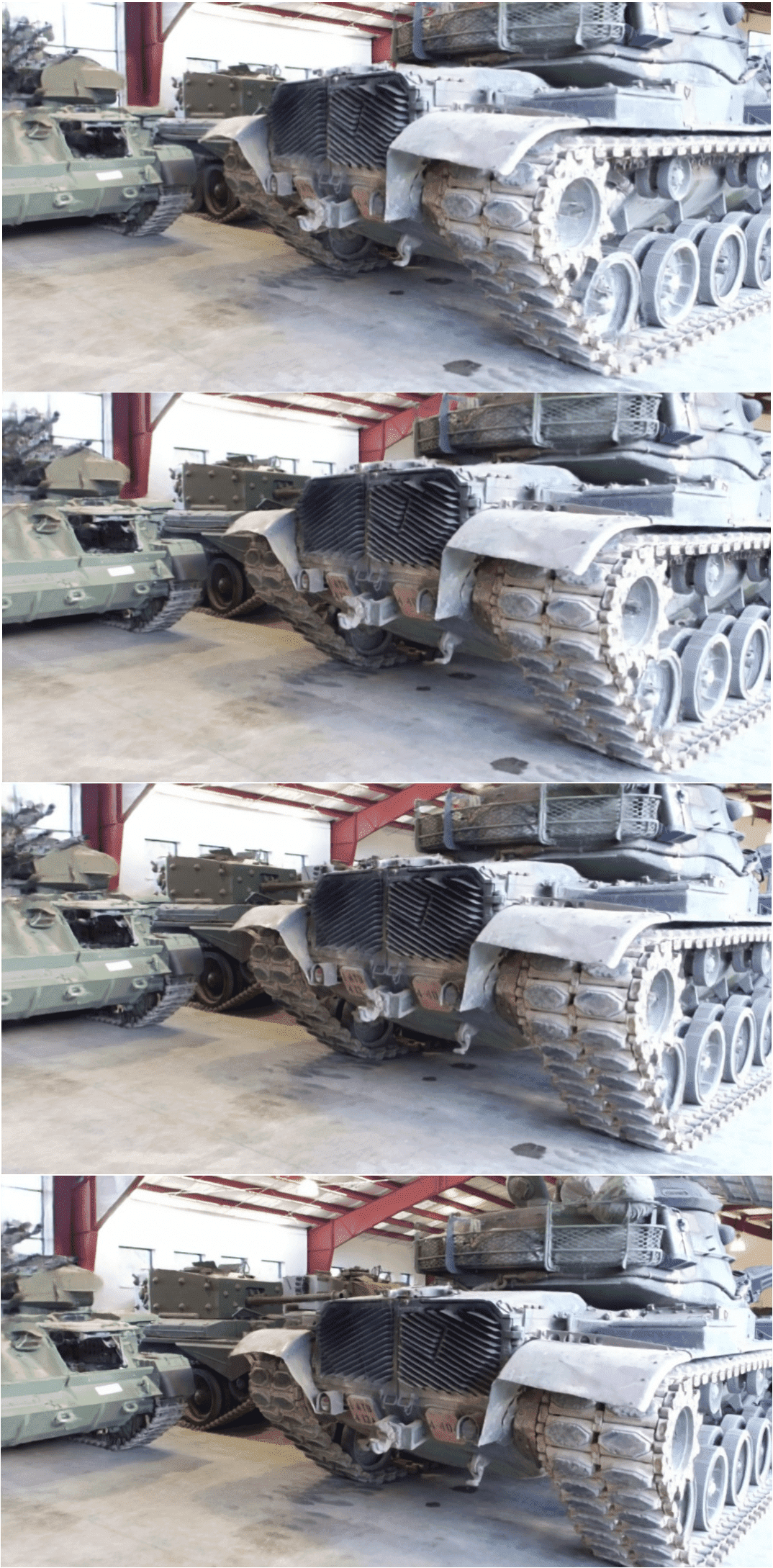}
     \end{subfigure}
\caption{Additional results on ``M60'' scene of the Tanks and Temples \cite{knapitsch2017tanks} dataset.}     
 \label{fig:qualitative_T_T_2}
\end{figure*}

\begin{figure*}[h!]
     \centering
     \begin{subfigure}[b]{0.2441\linewidth}
         \centering
         \includegraphics[width=\linewidth]{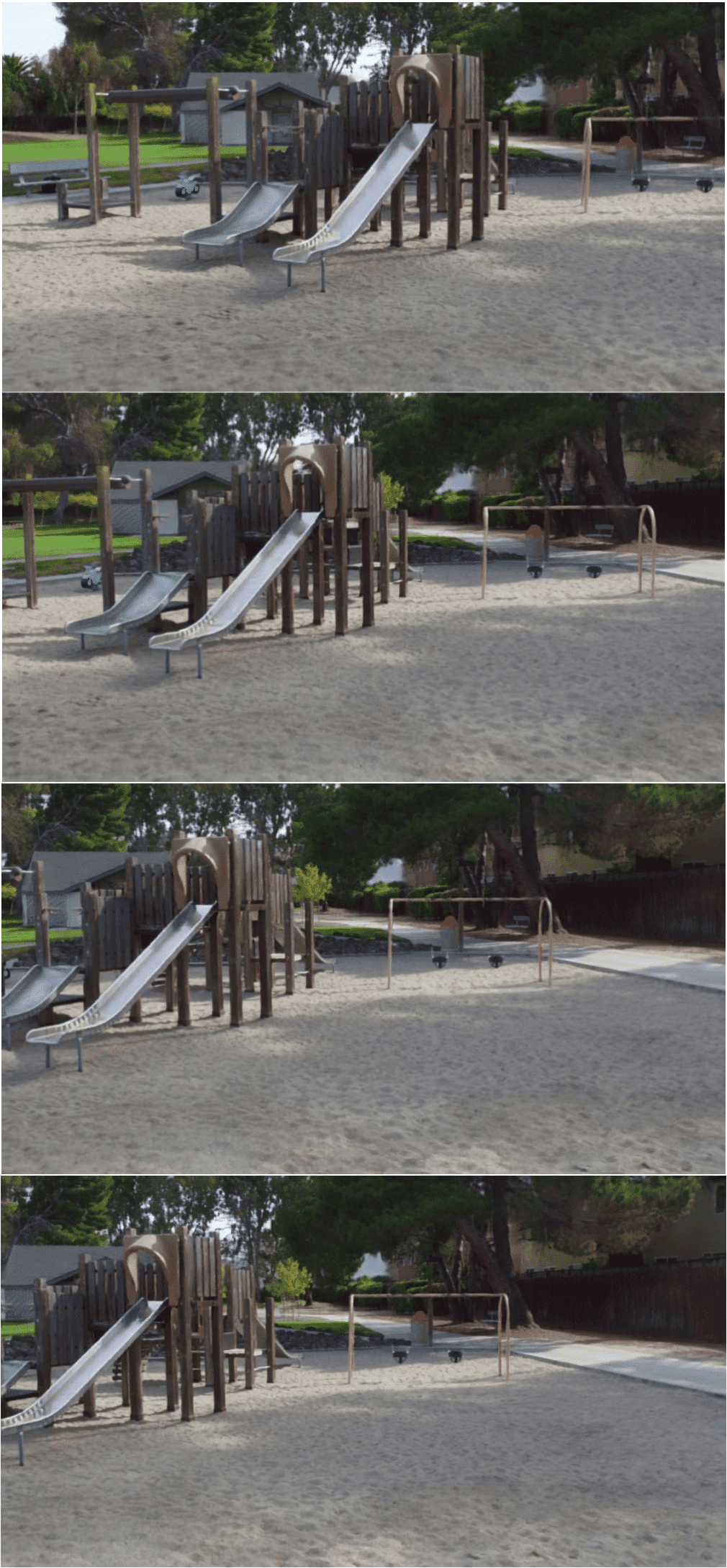}
     \end{subfigure}
     \hfill
     \begin{subfigure}[b]{0.2441\linewidth}
         \centering
         \includegraphics[width=\linewidth]{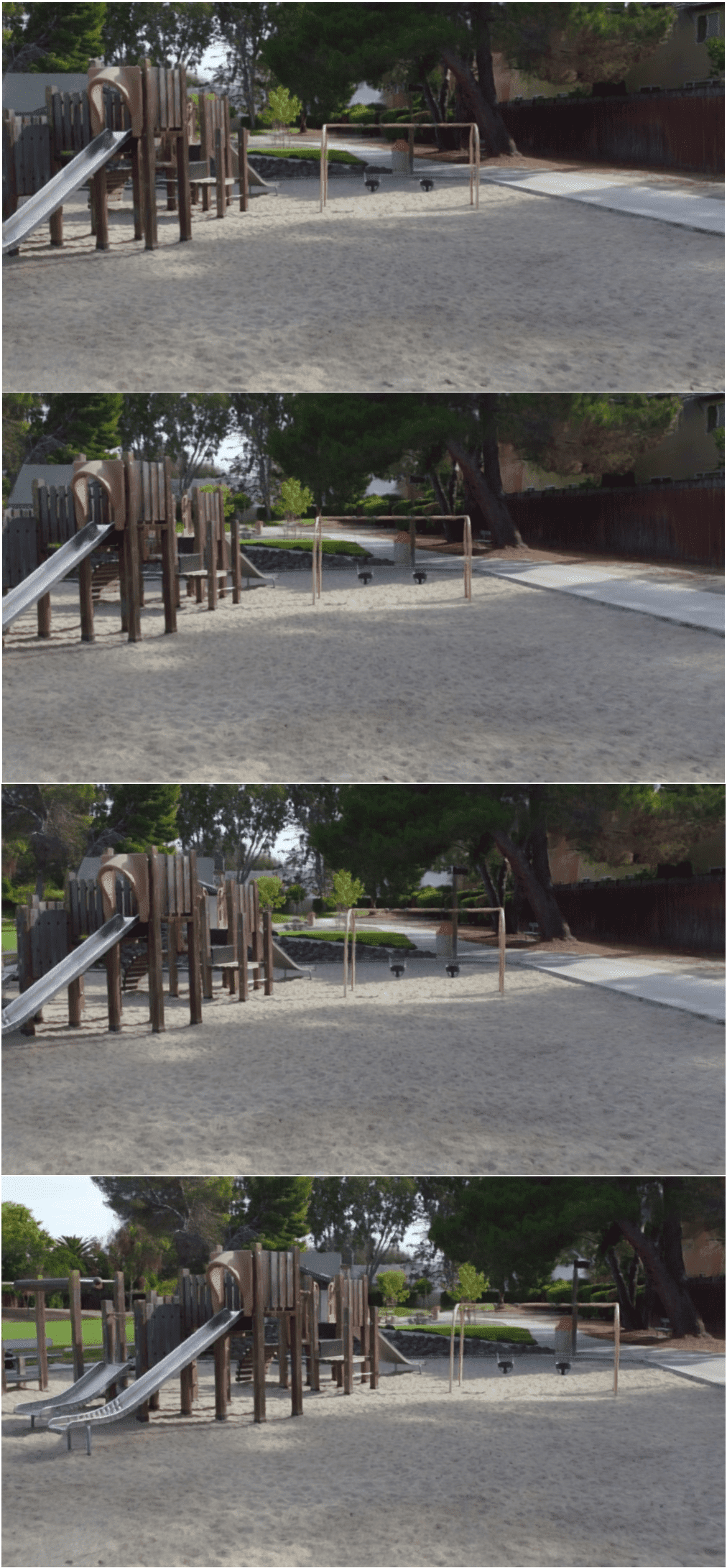}
     \end{subfigure}
      \hfill
     \begin{subfigure}[b]{0.2441\linewidth}
         \centering
         \includegraphics[width=\linewidth]{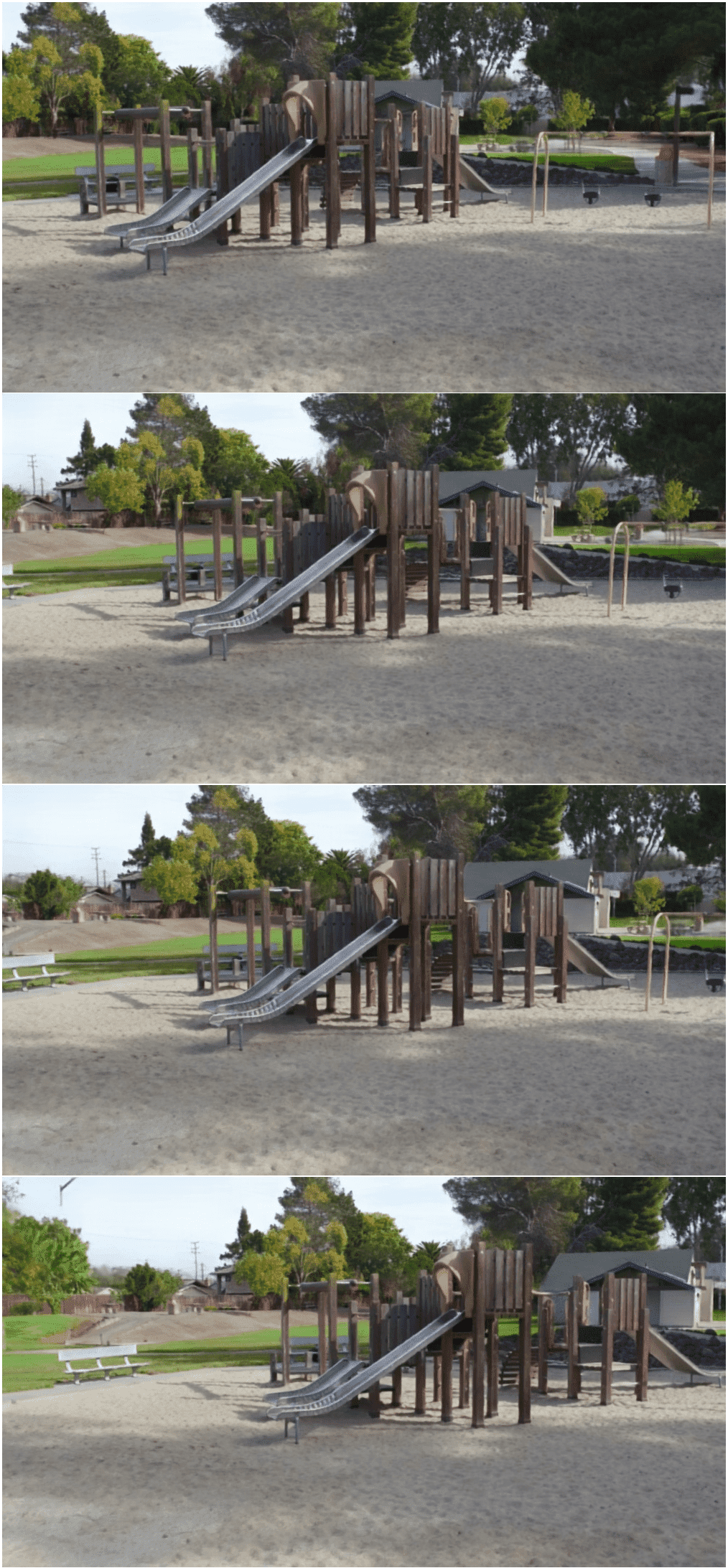}
     \end{subfigure}
      \hfill
     \begin{subfigure}[b]{0.2441\linewidth}
         \centering
         \includegraphics[width=\linewidth]{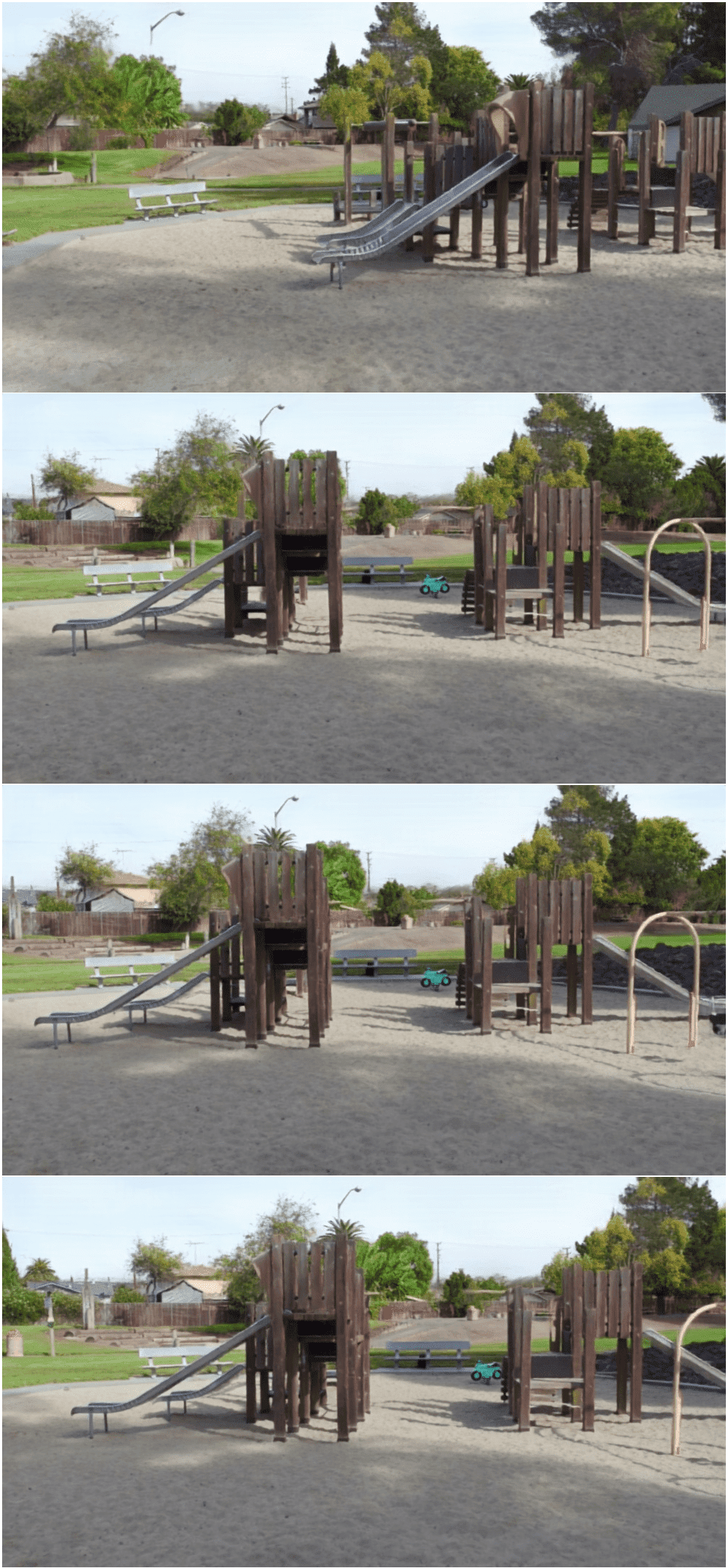}
     \end{subfigure}
\caption{Additional results on ``Playground'' scene of the Tanks and Temples \cite{knapitsch2017tanks} dataset.}     
 \label{fig:qualitative_T_T_3}
\end{figure*}

\begin{figure*}[h!]
     \centering
     \begin{subfigure}[b]{0.2441\linewidth}
         \centering
         \includegraphics[width=\linewidth]{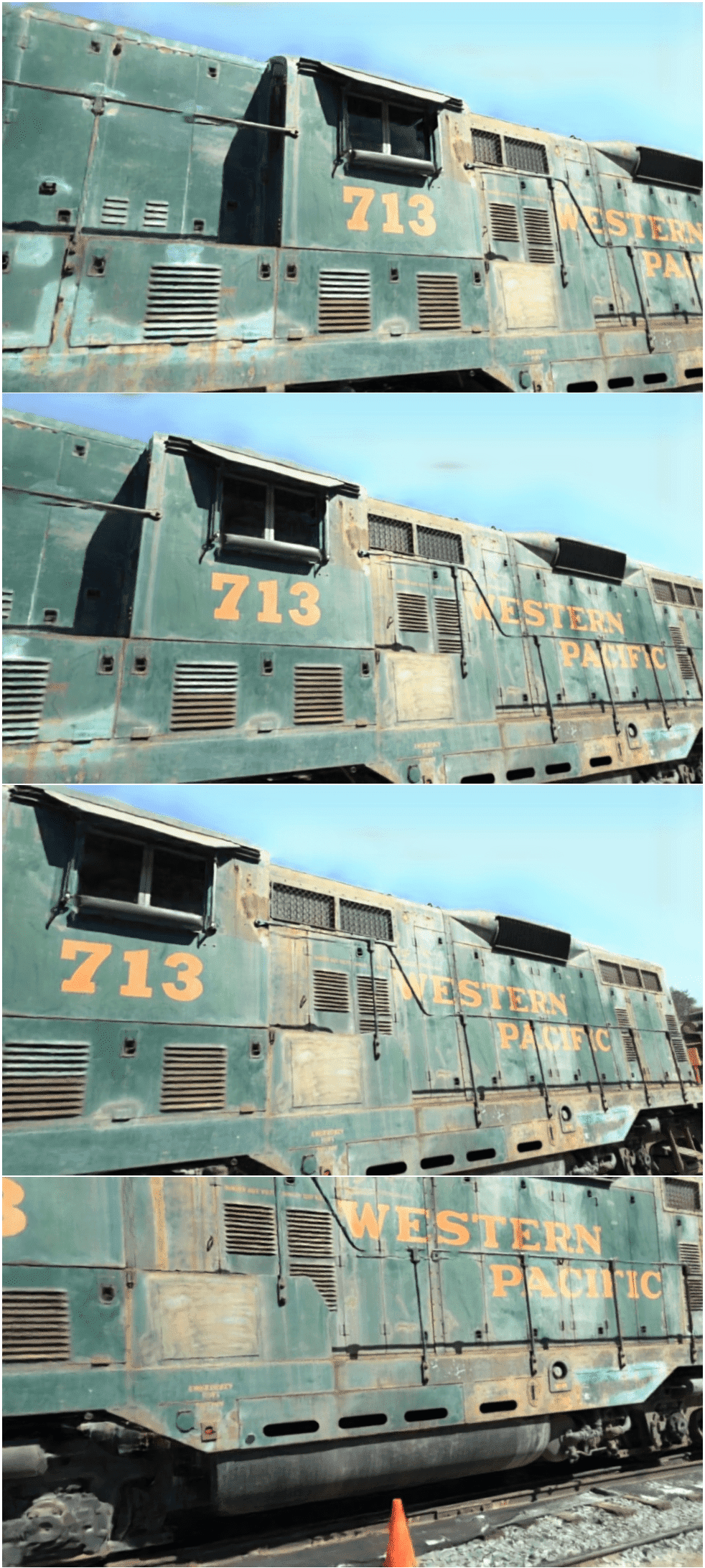}
     \end{subfigure}
     \hfill
     \begin{subfigure}[b]{0.2441\linewidth}
         \centering
         \includegraphics[width=\linewidth]{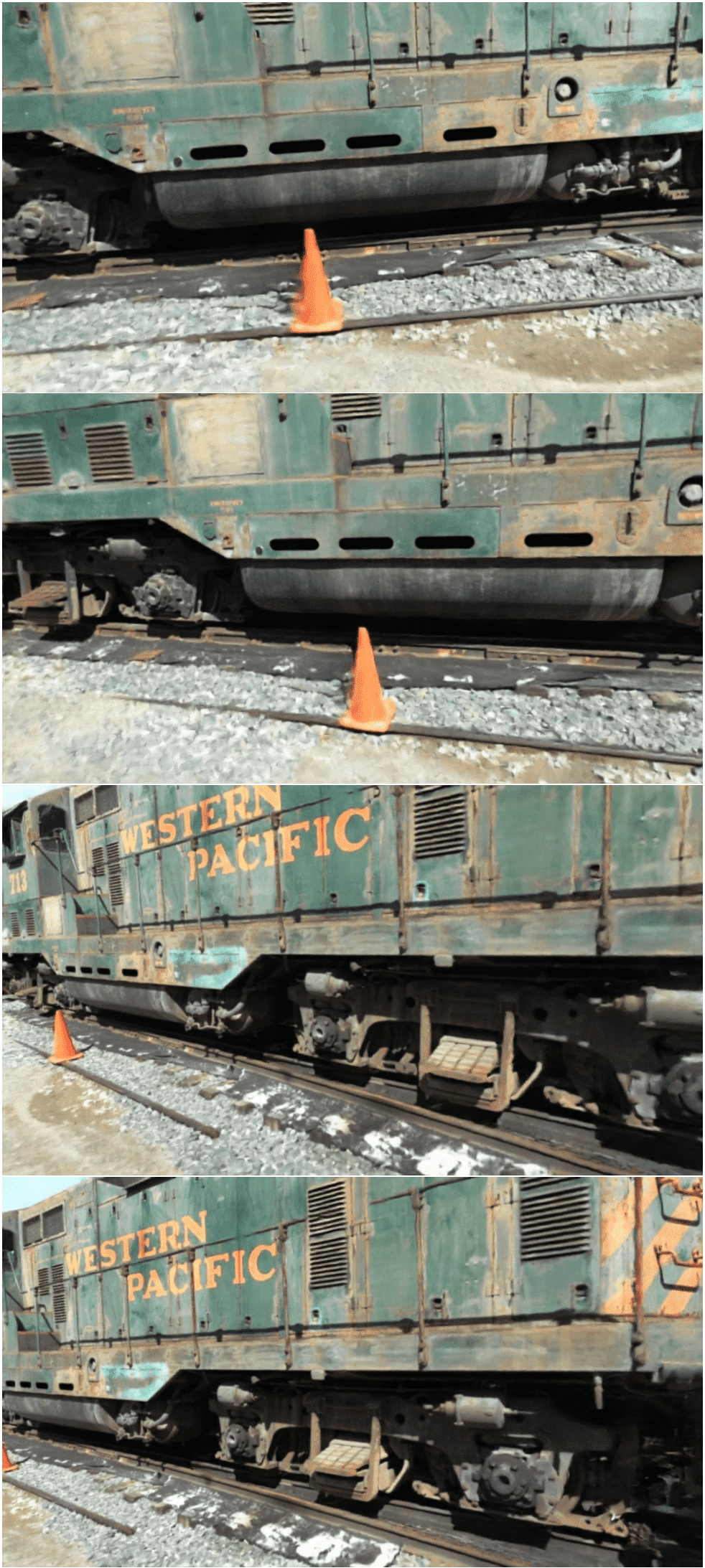}
     \end{subfigure}
      \hfill
     \begin{subfigure}[b]{0.2441\linewidth}
         \centering
         \includegraphics[width=\linewidth]{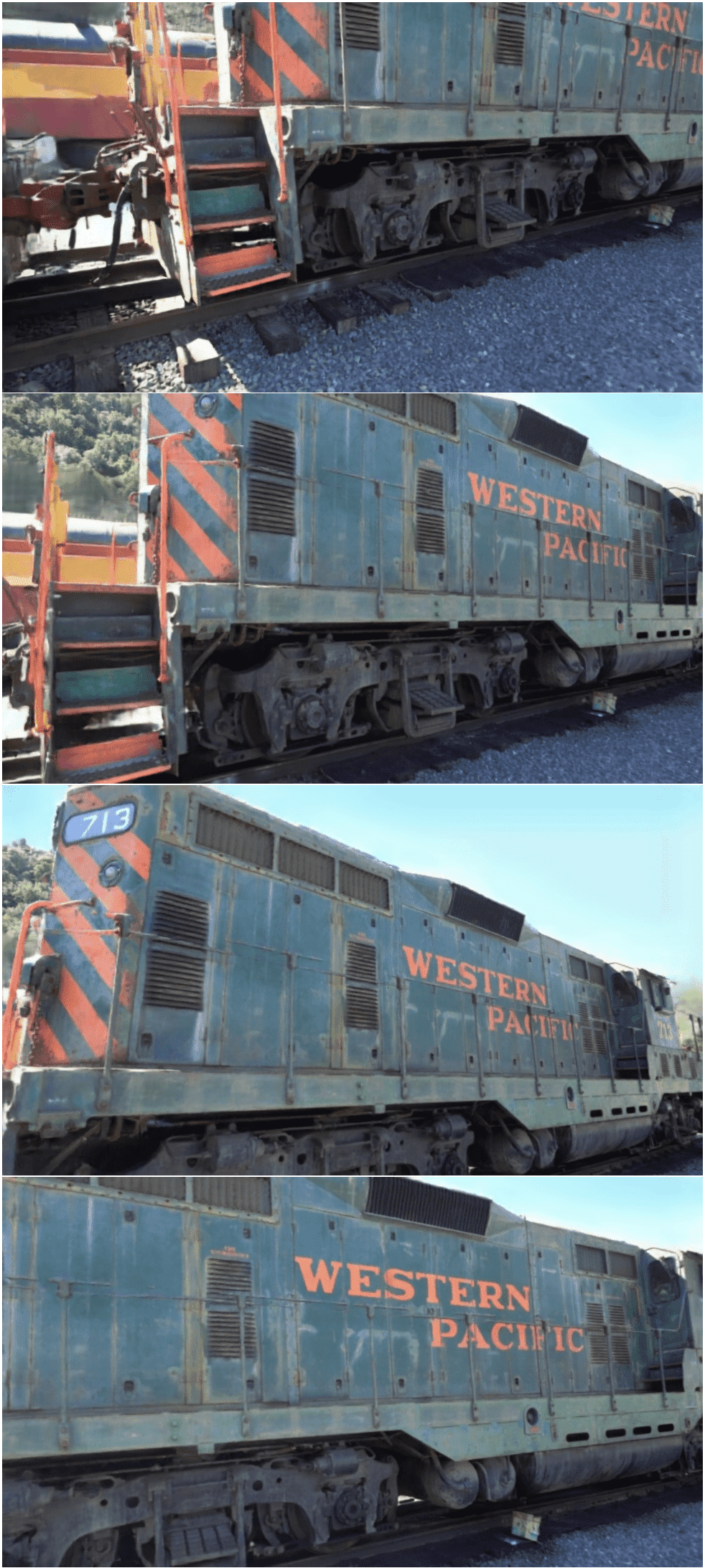}
     \end{subfigure}
      \hfill
     \begin{subfigure}[b]{0.2441\linewidth}
         \centering
         \includegraphics[width=\linewidth]{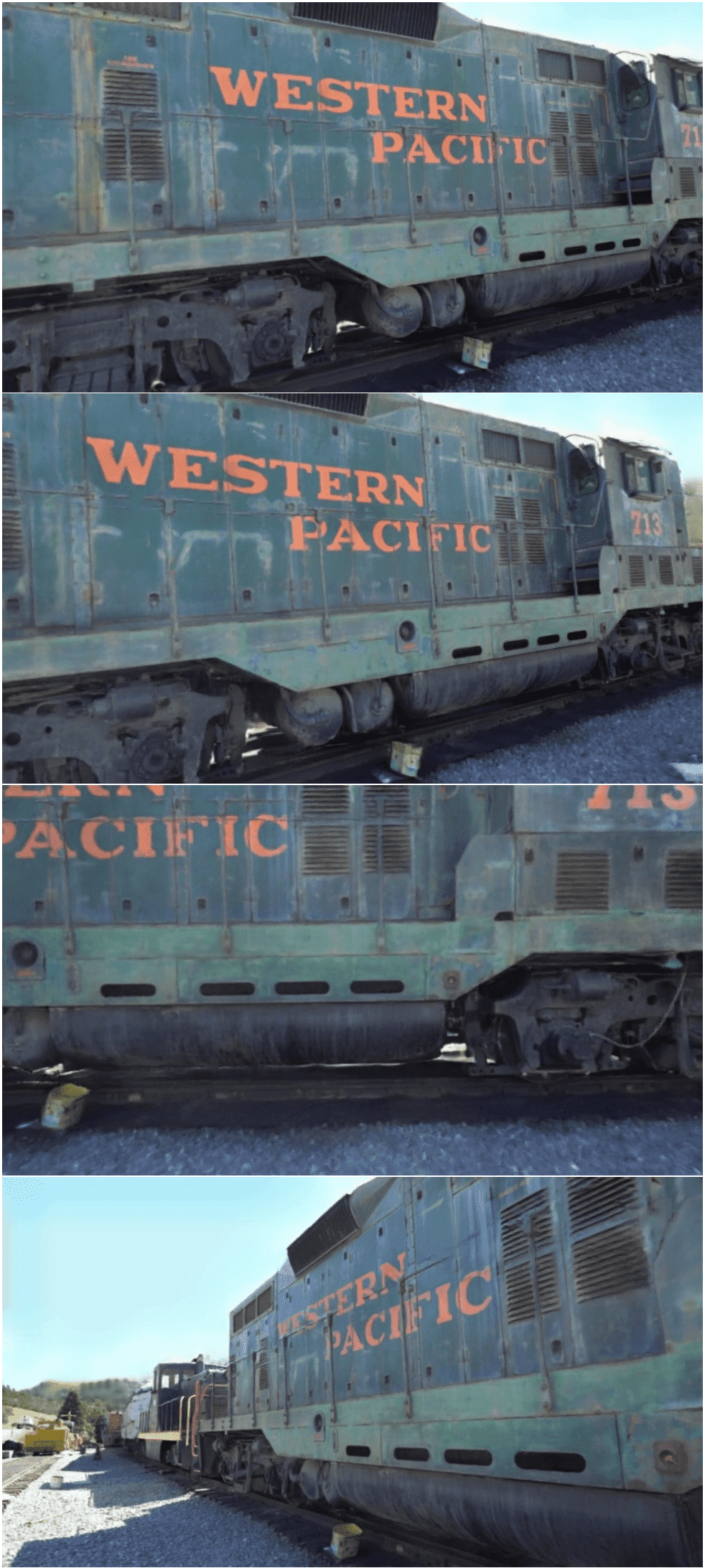}
     \end{subfigure}
\caption{Additional results on ``Train'' scene of the Tanks and Temples \cite{knapitsch2017tanks} dataset.}     
 \label{fig:qualitative_T_T_4}
\end{figure*}

\begin{figure*}[h!]
     \centering
     \begin{subfigure}[t]{0.22\linewidth}
         \centering
         \includegraphics[width=\linewidth]{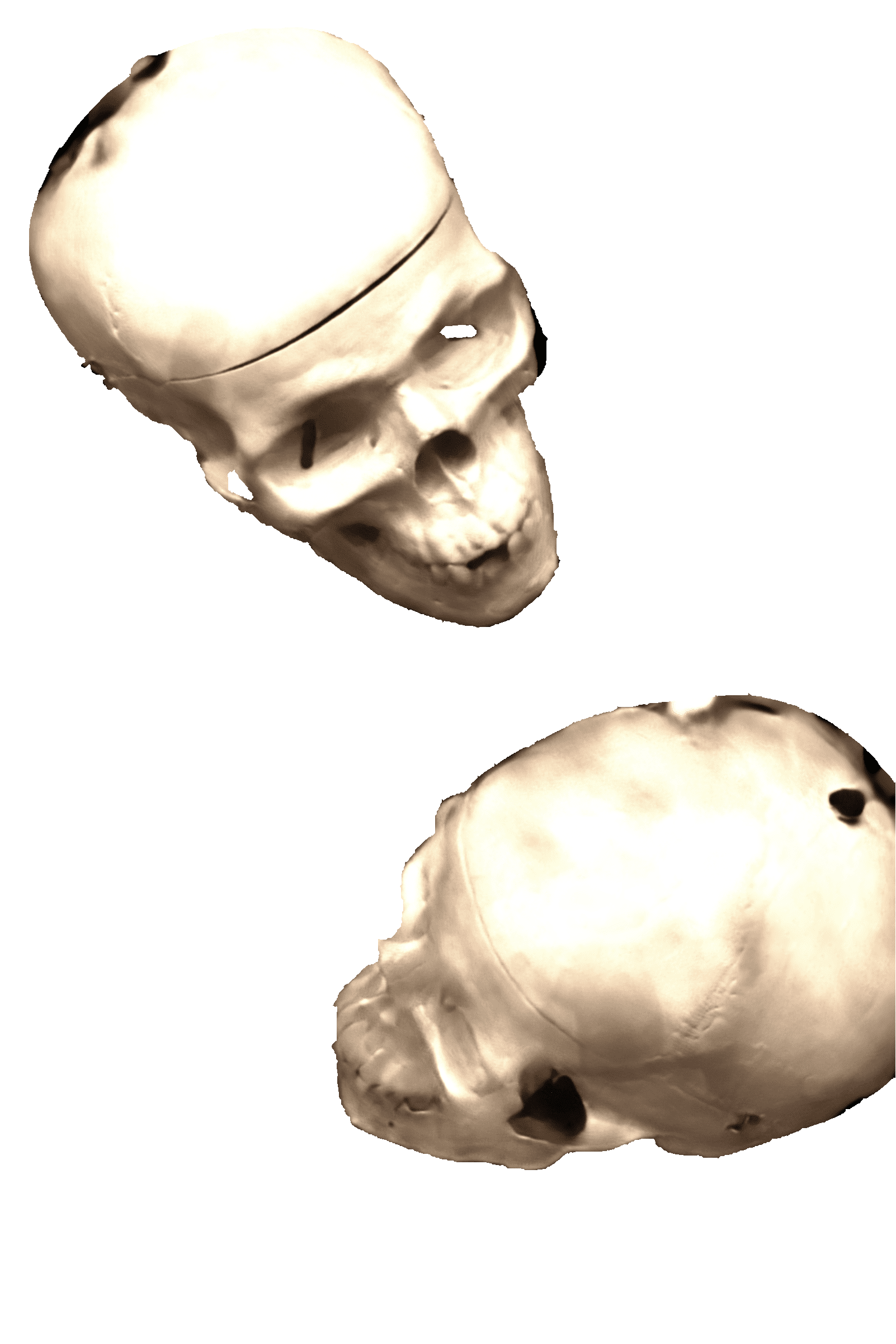}
     \end{subfigure}
     \begin{subfigure}[t]{0.22\linewidth}
         \centering
         \includegraphics[width=\linewidth]{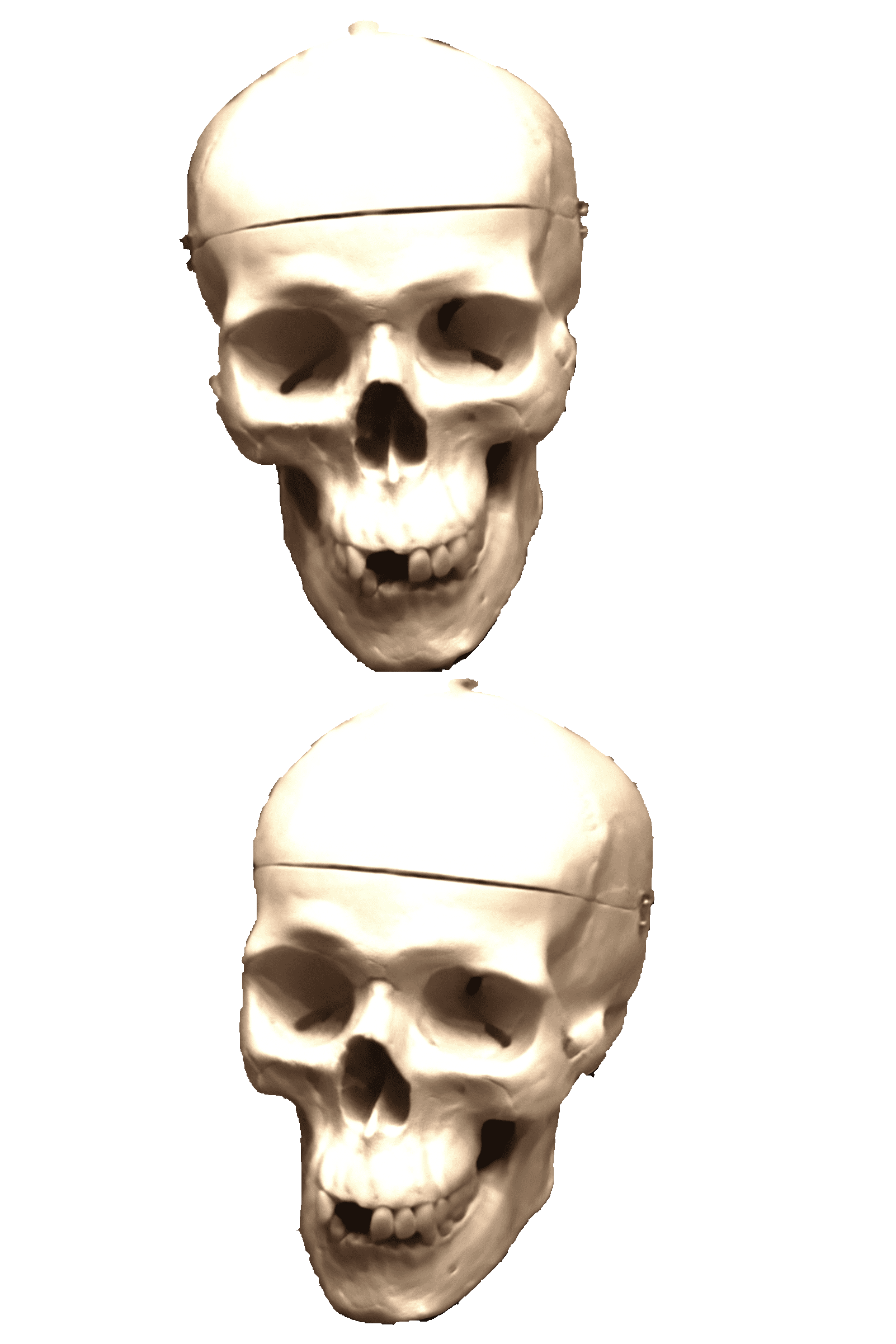}
     \end{subfigure}
     \begin{subfigure}[t]{0.22\linewidth}
         \centering
         \includegraphics[width=\linewidth]{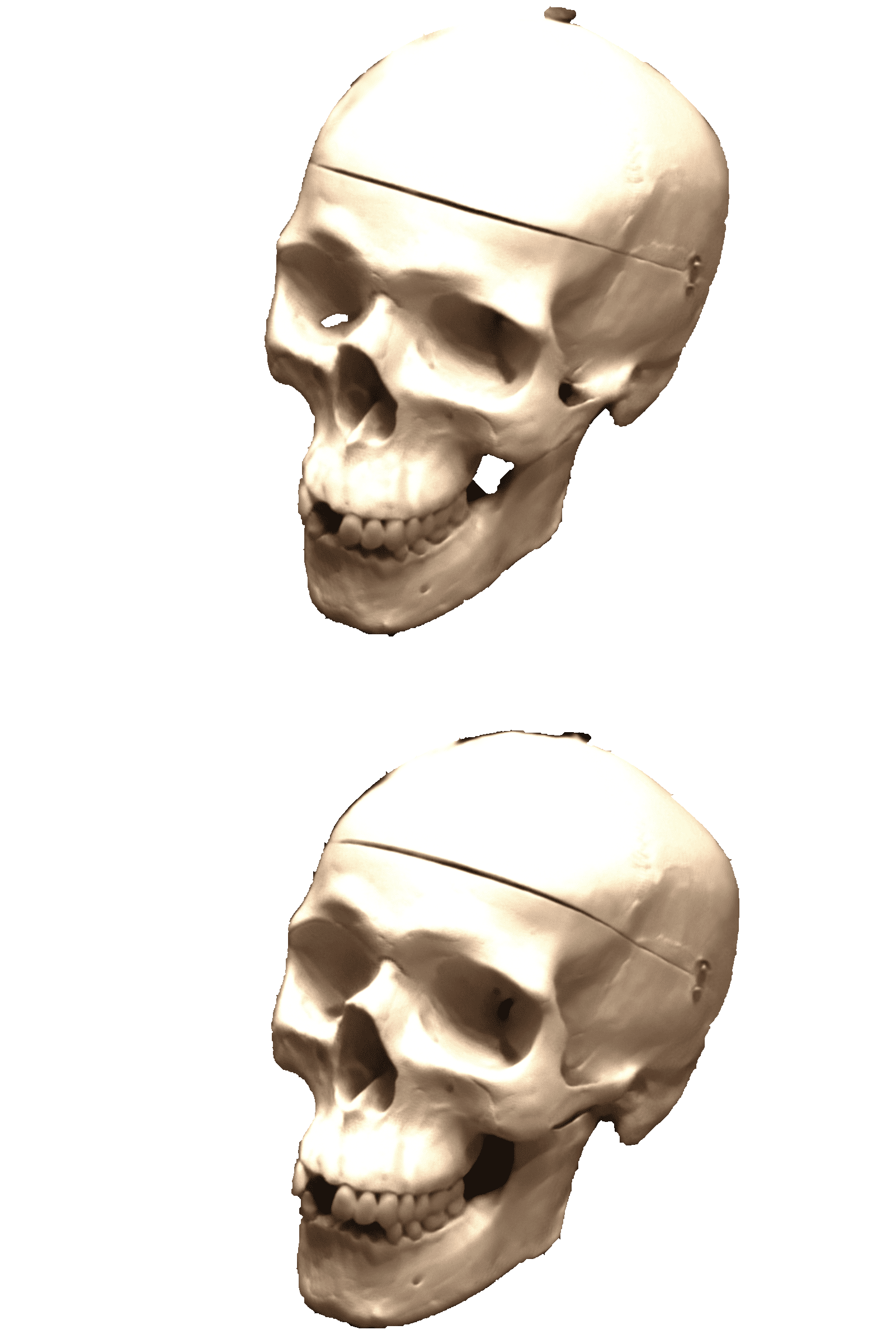}
     \end{subfigure}
     \begin{subfigure}[t]{0.22\linewidth}
         \centering
         \includegraphics[width=\linewidth]{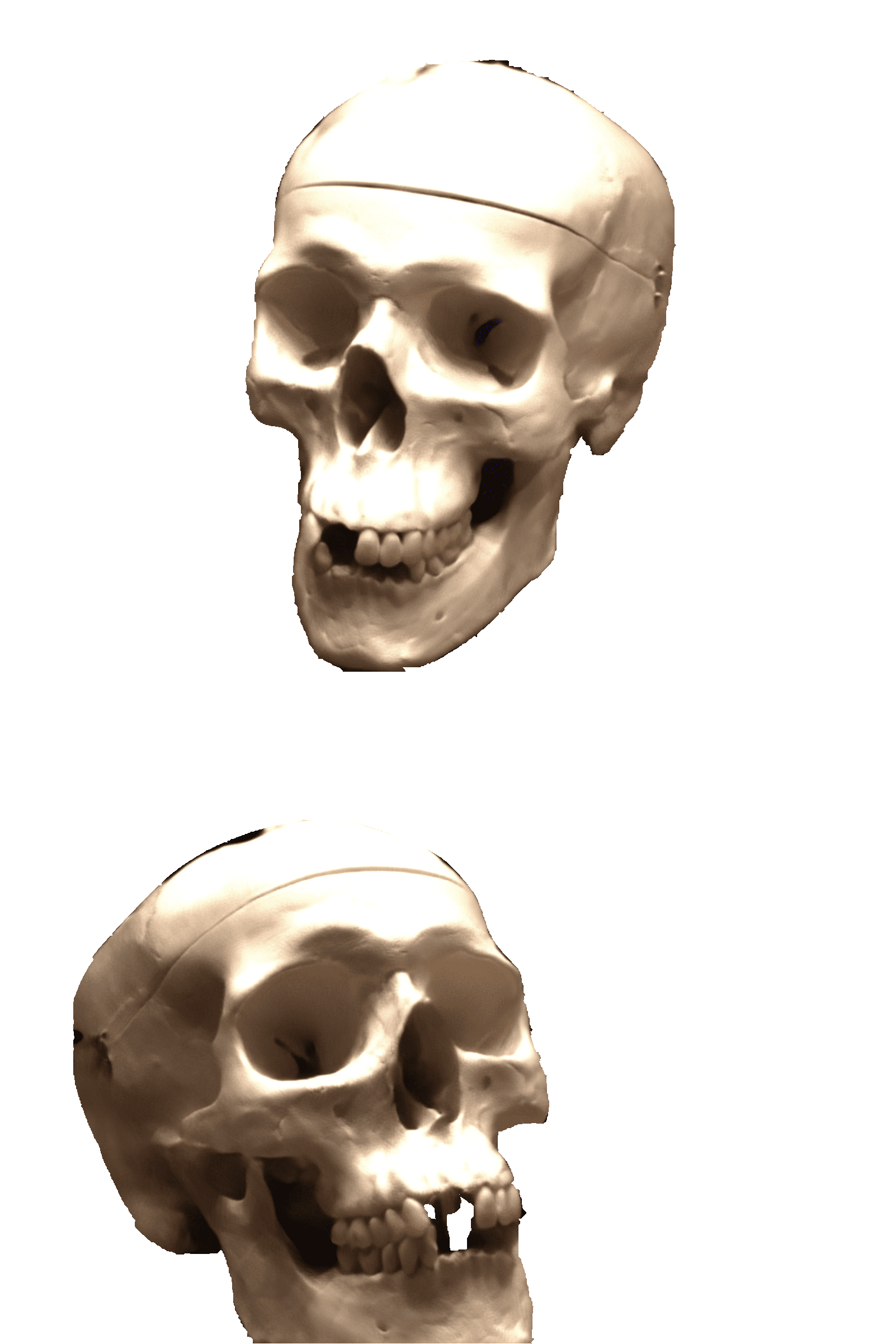}
     \end{subfigure}
 \caption{Additional results on scene ``65'' of the DTU~\cite{aanaes2016large} dataset.}     
 \label{fig:qualitative_DTU_65}
\end{figure*}

\begin{figure*}[h!]
     \centering
     \begin{subfigure}[t]{0.22\linewidth}
         \centering
         \includegraphics[width=\linewidth]{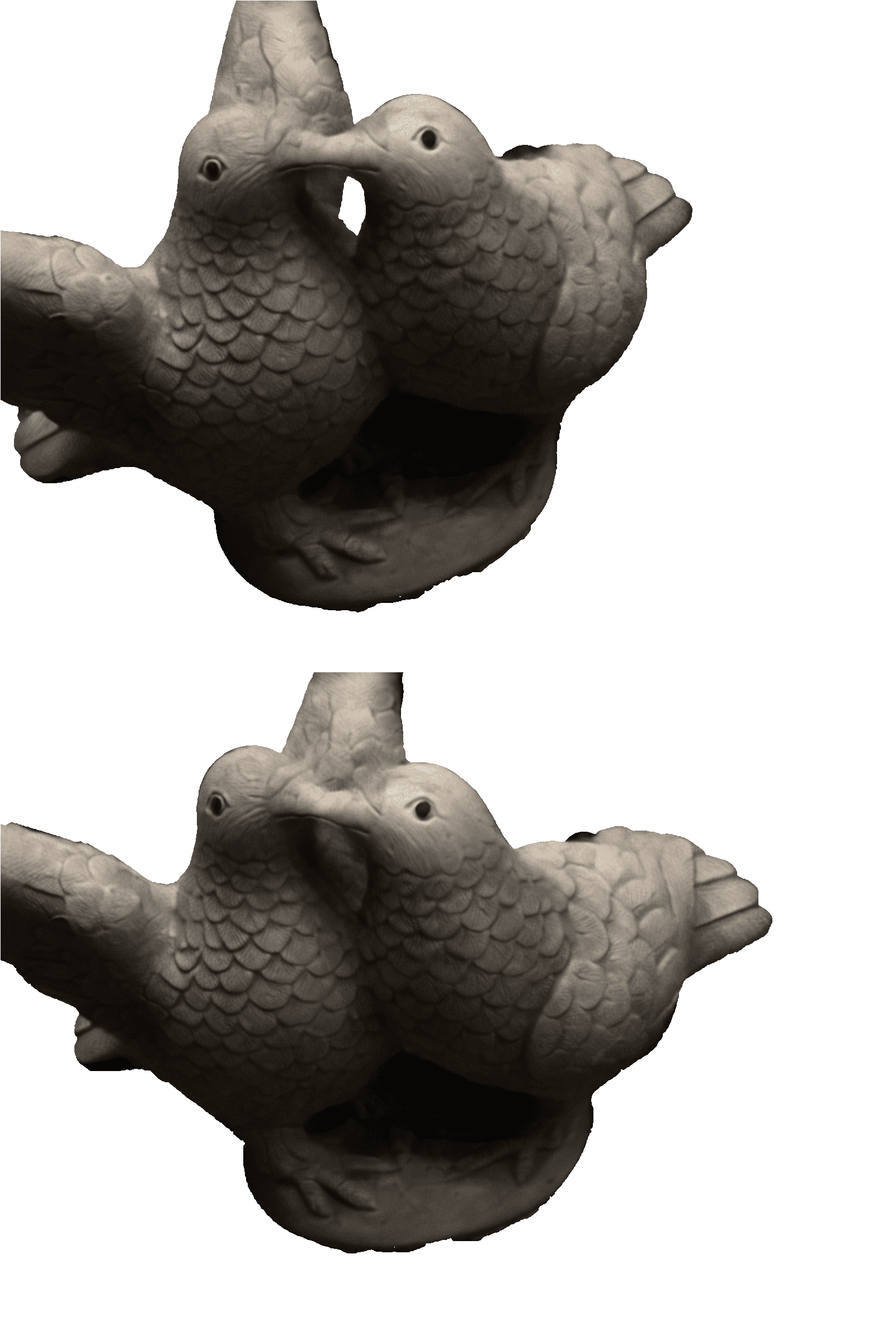}
     \end{subfigure}
     \begin{subfigure}[t]{0.22\linewidth}
         \centering
         \includegraphics[width=\linewidth]{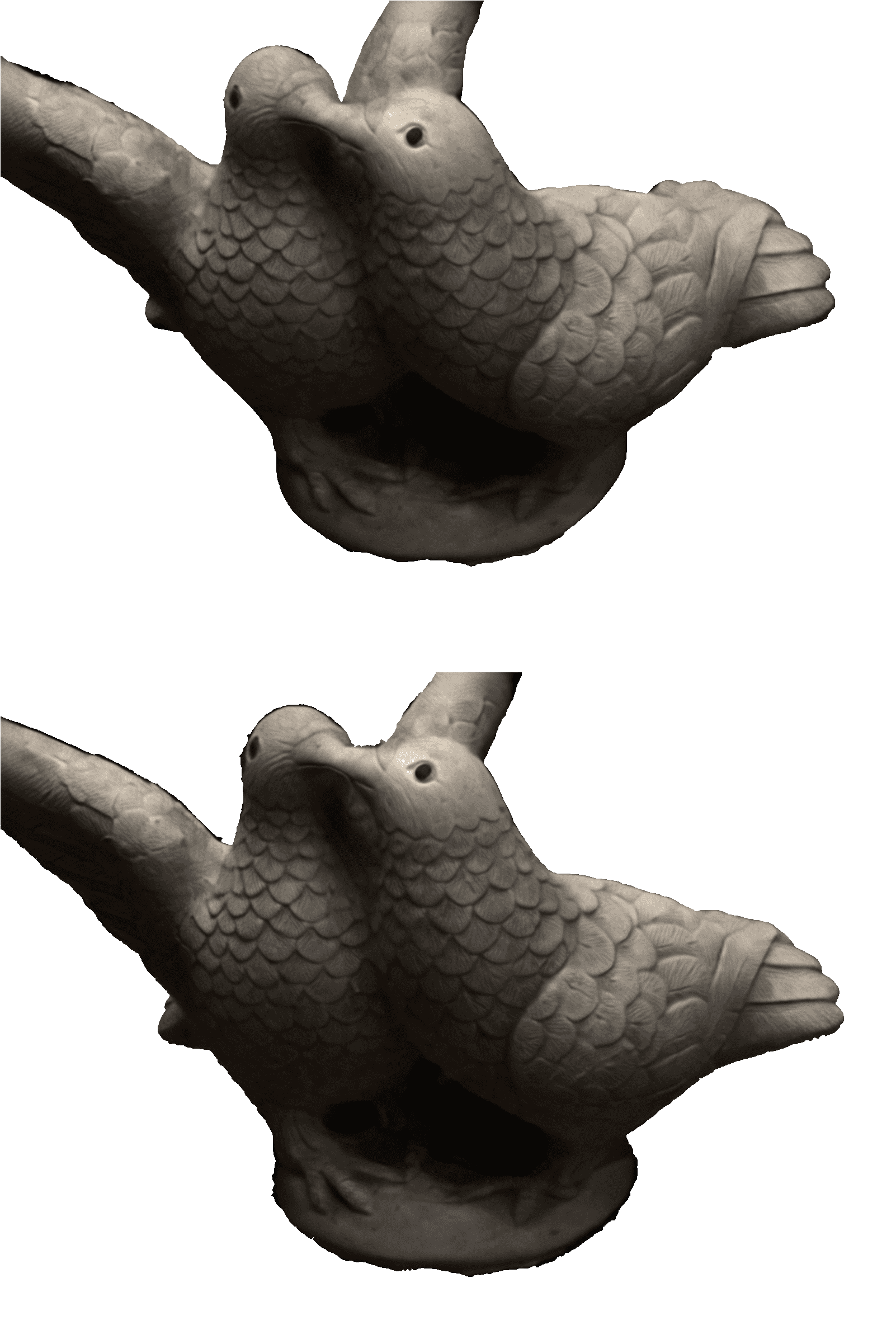}
     \end{subfigure}
     \begin{subfigure}[t]{0.22\linewidth}
         \centering
         \includegraphics[width=\linewidth]{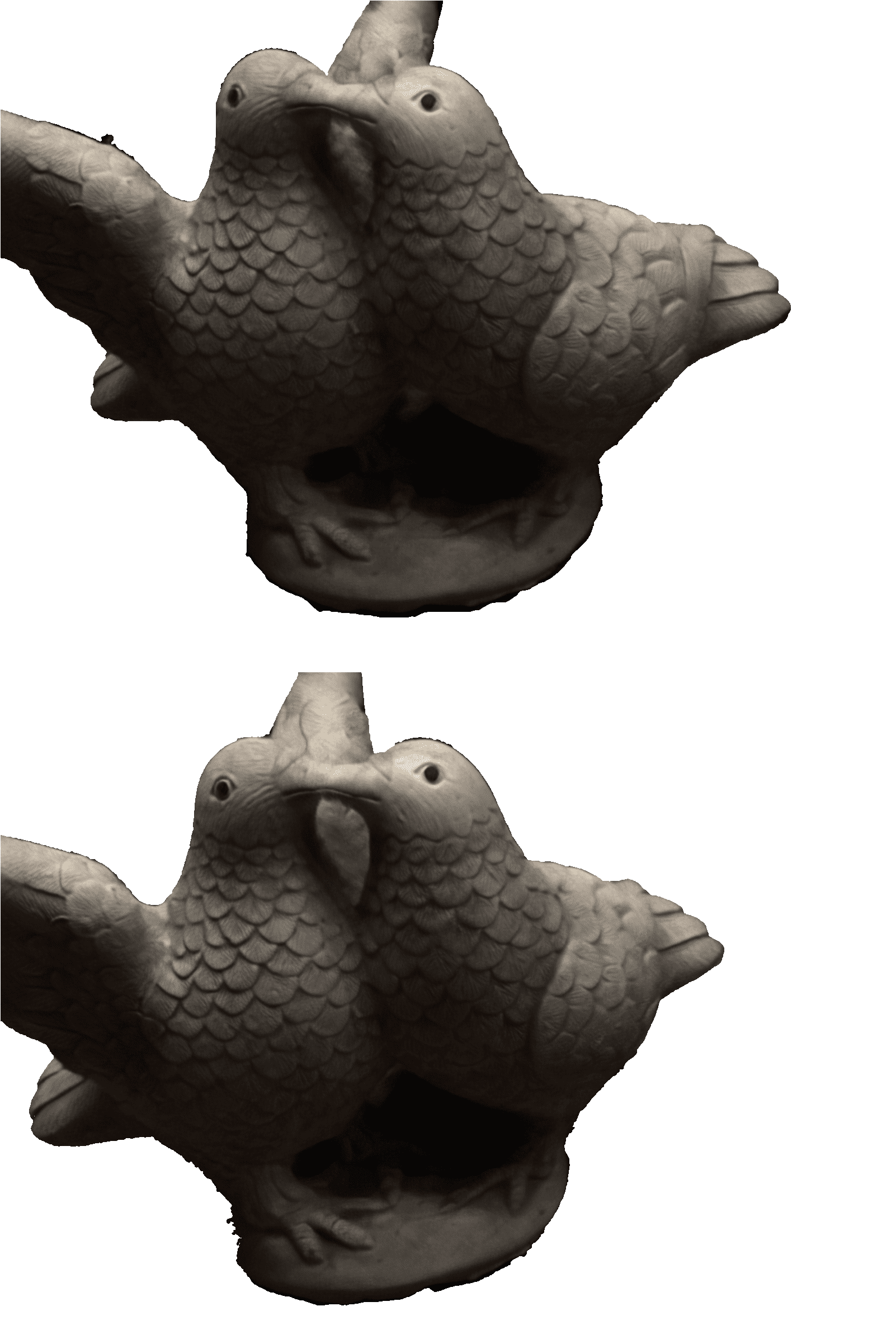}
     \end{subfigure}
     \begin{subfigure}[t]{0.22\linewidth}
         \centering
         \includegraphics[width=\linewidth]{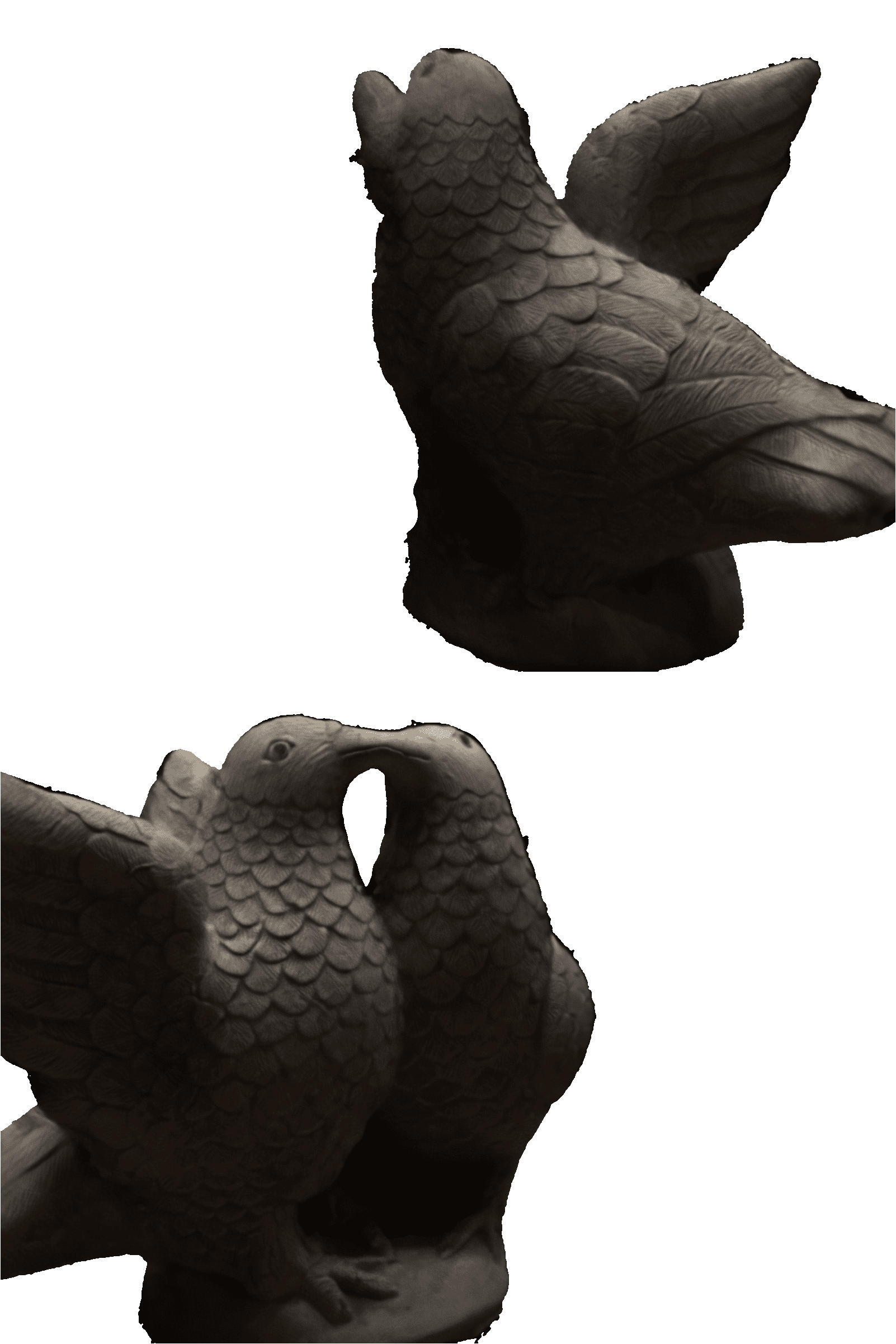}
     \end{subfigure}
 \caption{Additional results on scene ``106'' of the DTU~\cite{aanaes2016large} dataset.}     
 \label{fig:qualitative_DTU_106}
\end{figure*}

\begin{figure*}[h!]
     \centering
     \begin{subfigure}[t]{0.22\linewidth}
         \centering
         \includegraphics[width=\linewidth]{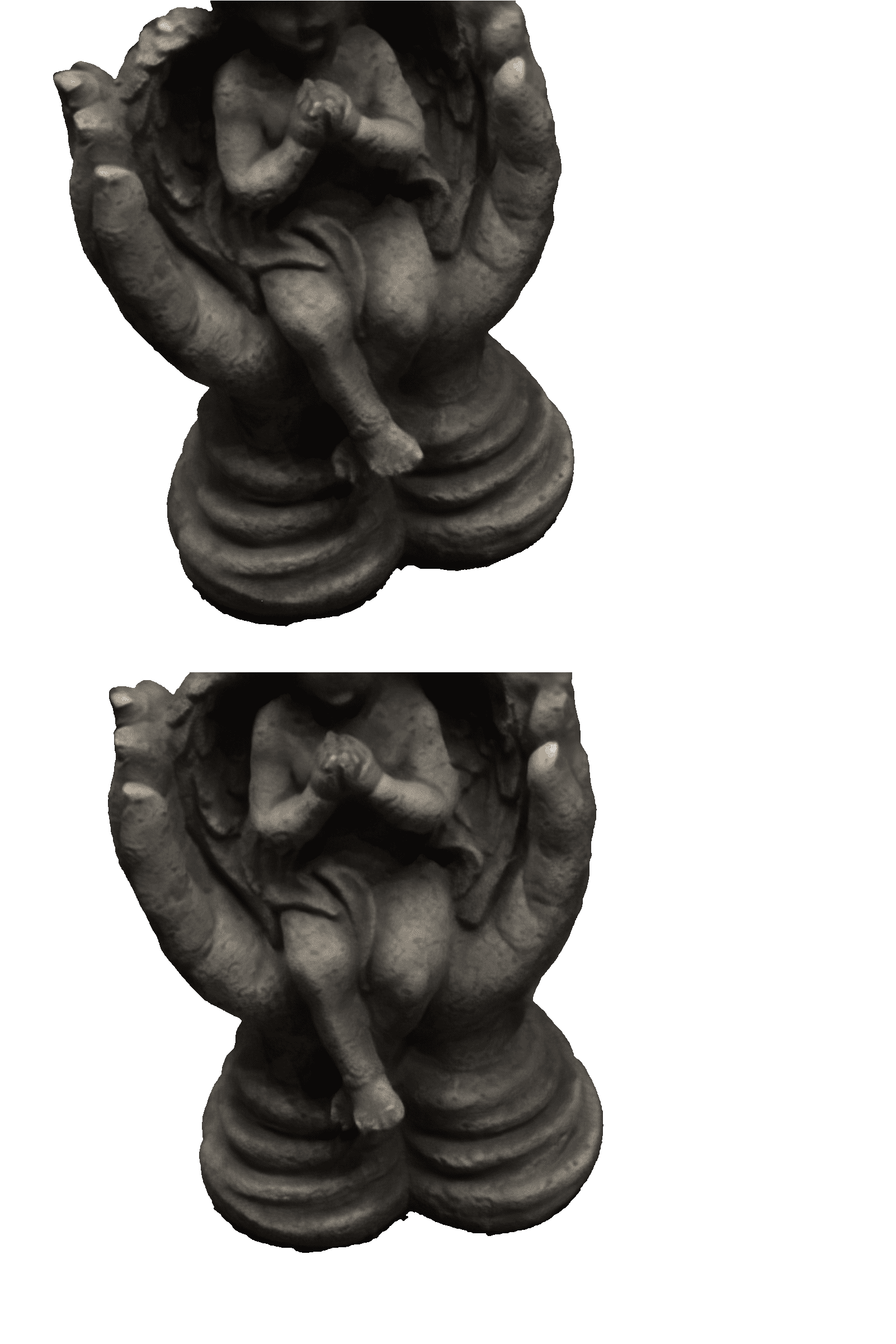}
     \end{subfigure}
     \begin{subfigure}[t]{0.22\linewidth}
         \centering
         \includegraphics[width=\linewidth]{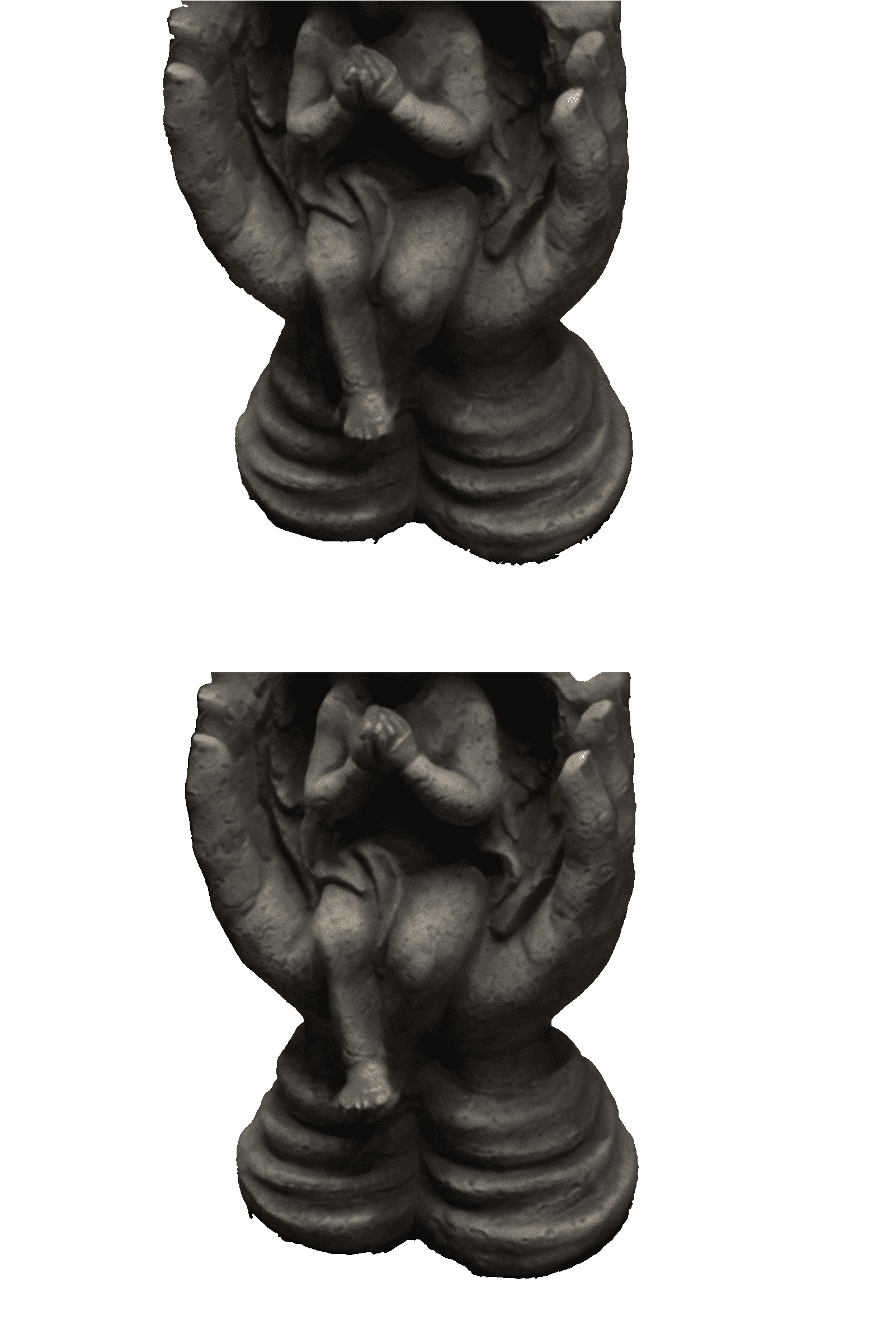}
     \end{subfigure}
     \begin{subfigure}[t]{0.22\linewidth}
         \centering
         \includegraphics[width=\linewidth]{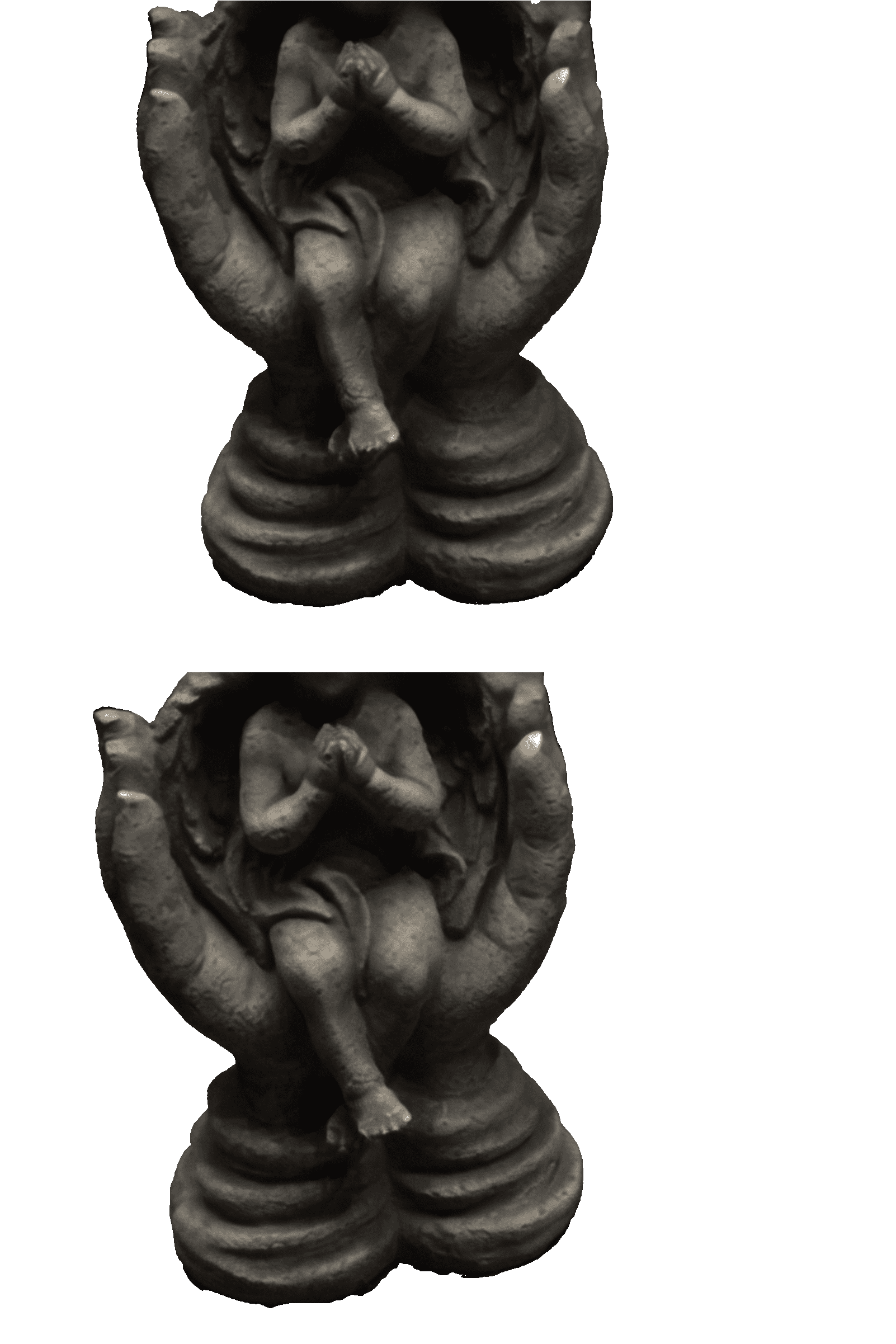}
     \end{subfigure}
     \begin{subfigure}[t]{0.22\linewidth}
         \centering
         \includegraphics[width=\linewidth]{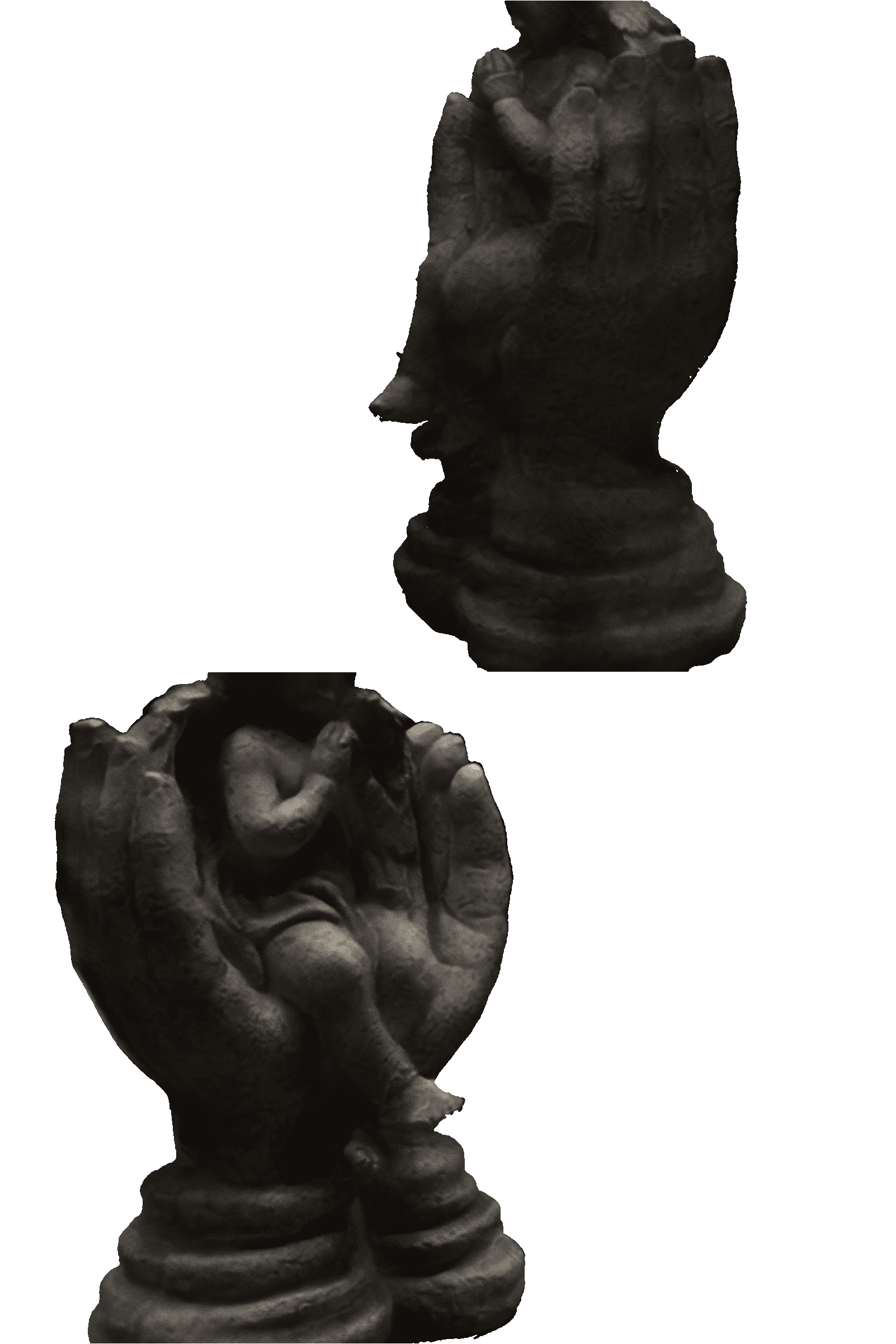}
     \end{subfigure}
 \caption{Additional results on scene ``118'' of the DTU~\cite{aanaes2016large} dataset.}     
 \label{fig:qualitative_DTU_118}
\end{figure*}

\paragraph{Results on Tanks and Temples~\cite{knapitsch2017tanks}:} Figures \ref{fig:qualitative_T_T_1}, \ref{fig:qualitative_T_T_2}, \ref{fig:qualitative_T_T_3}, \ref{fig:qualitative_T_T_4} show additional novel view synthesis qualitative results of our method for more test views on the Tanks and Temples~\cite{knapitsch2017tanks} dataset, not seen during training.
\paragraph{Results on DTU~\cite{aanaes2016large}:} Figures \ref{fig:qualitative_DTU_65}, \ref{fig:qualitative_DTU_106}, \ref{fig:qualitative_DTU_118} show additional novel view synthesis qualitative results of our method for more test views on the DTU~\cite{aanaes2016large} dataset, not seen during training.

\section{Additional qualitative comparisons}
\begin{figure*}[h!]
     \centering
     \begin{subfigure}[b]{0.2441\linewidth}
         \centering
         \includegraphics[width=\linewidth]{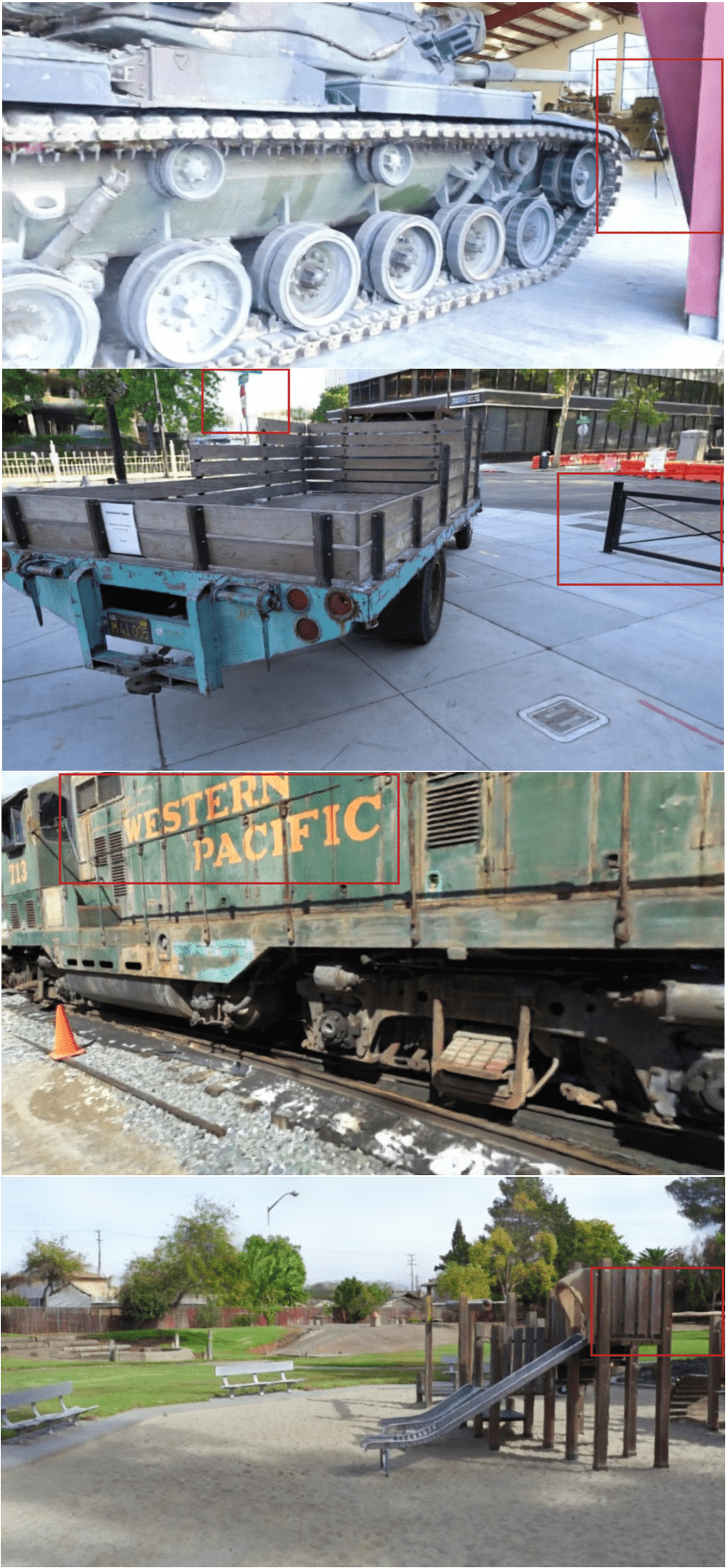}
        \caption{Ours}
     \end{subfigure}
     \hfill
     \begin{subfigure}[b]{0.2441\linewidth}
         \centering
         \includegraphics[width=\linewidth]{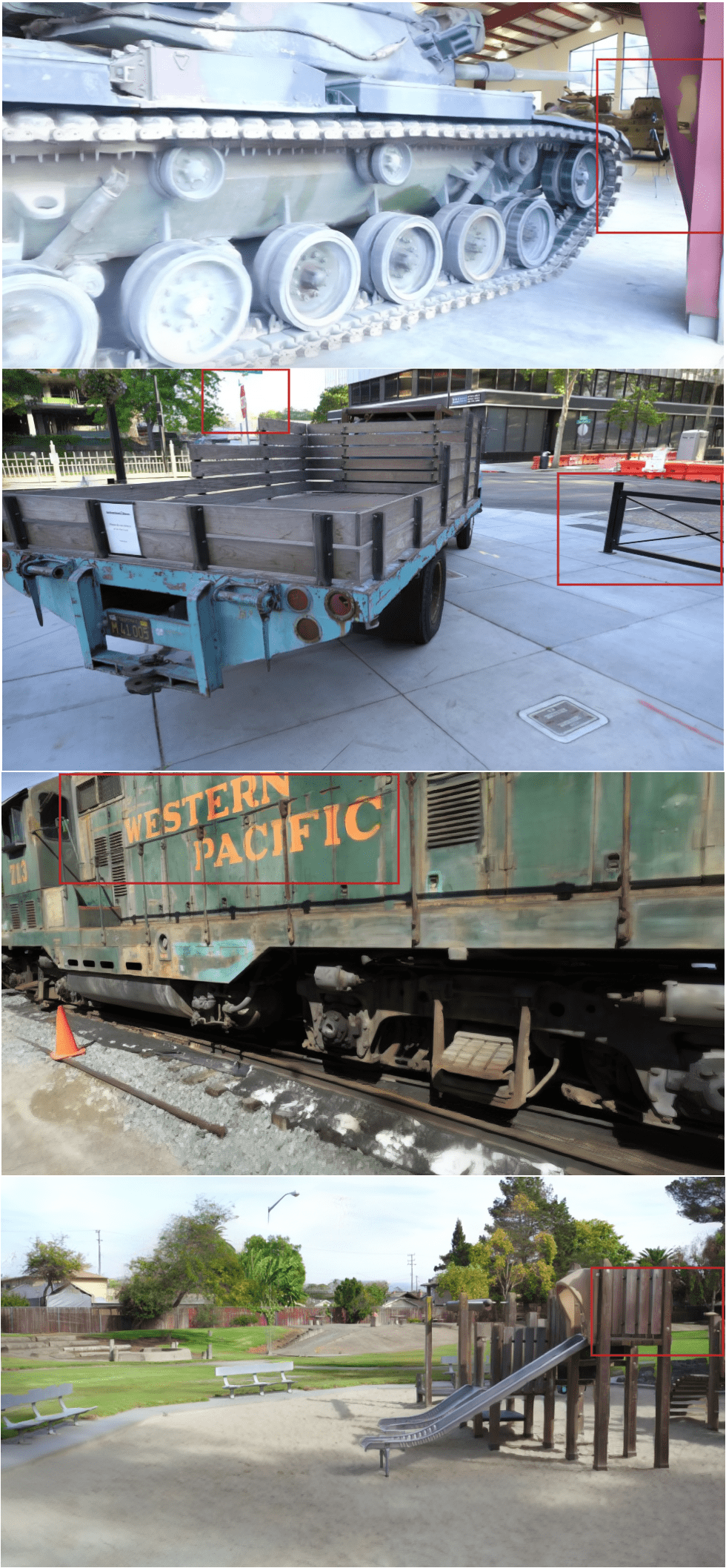}
        \caption{SVS~\cite{riegler2021stable}}
     \end{subfigure}
      \hfill
     \begin{subfigure}[b]{0.2441\linewidth}
         \centering
         \includegraphics[width=\linewidth]{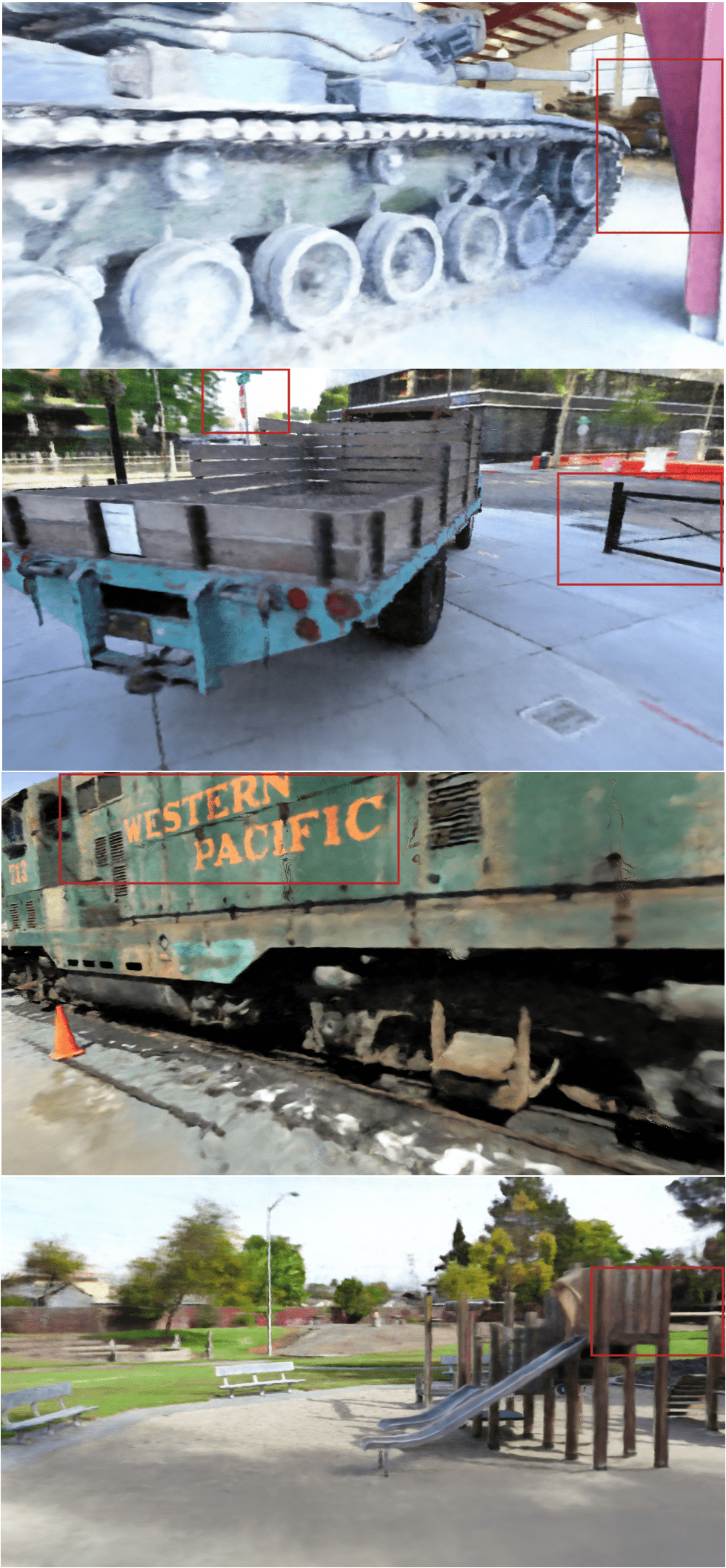}
        \caption{NeRF++~\cite{zhang2020nerf++}}
     \end{subfigure}
      \hfill
     \begin{subfigure}[b]{0.2441\linewidth}
         \centering
         \includegraphics[width=\linewidth]{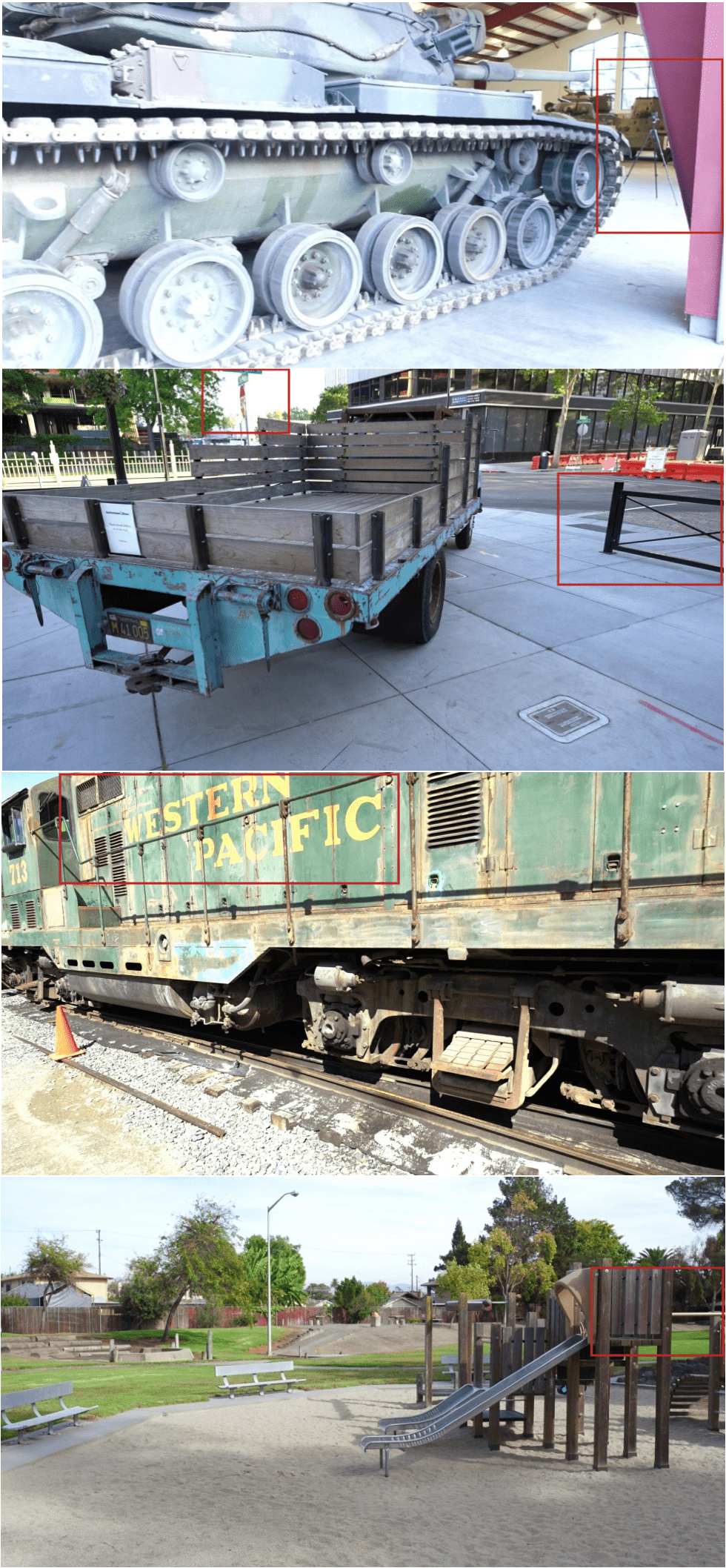}
        \caption{GT}
     \end{subfigure}
\caption{Additional qualitative comparisons on Tanks and Temples \cite{knapitsch2017tanks}.}     
 \label{fig:qualitative_T_T_compare}
\end{figure*}

\begin{figure*}[h!]
     \centering
     \begin{subfigure}[t]{0.22\linewidth}
         \centering
         \includegraphics[width=\linewidth]{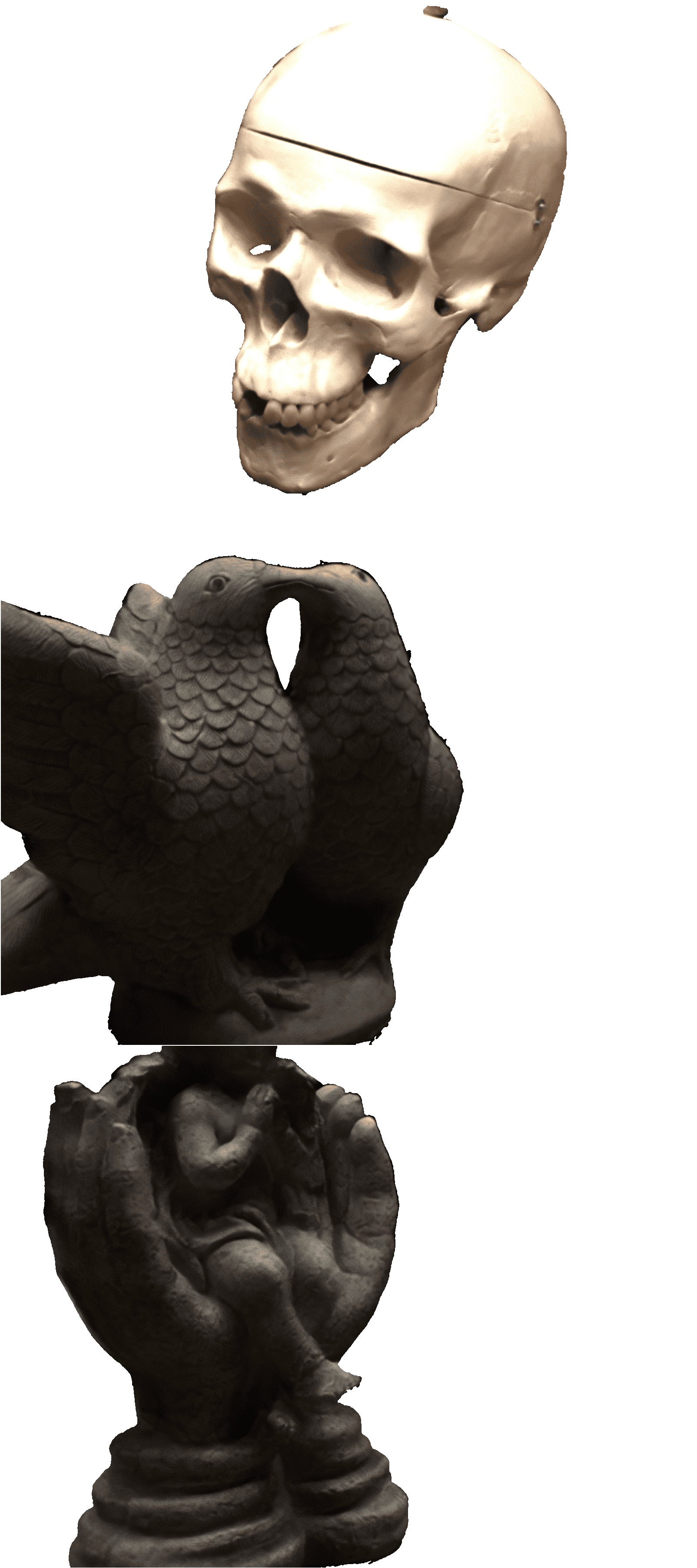}
         \captionsetup{font=tiny,labelfont=tiny}
        \caption{Ours (Full)}
     \end{subfigure}
     \begin{subfigure}[t]{0.22\linewidth}
         \centering
         \includegraphics[width=\linewidth]{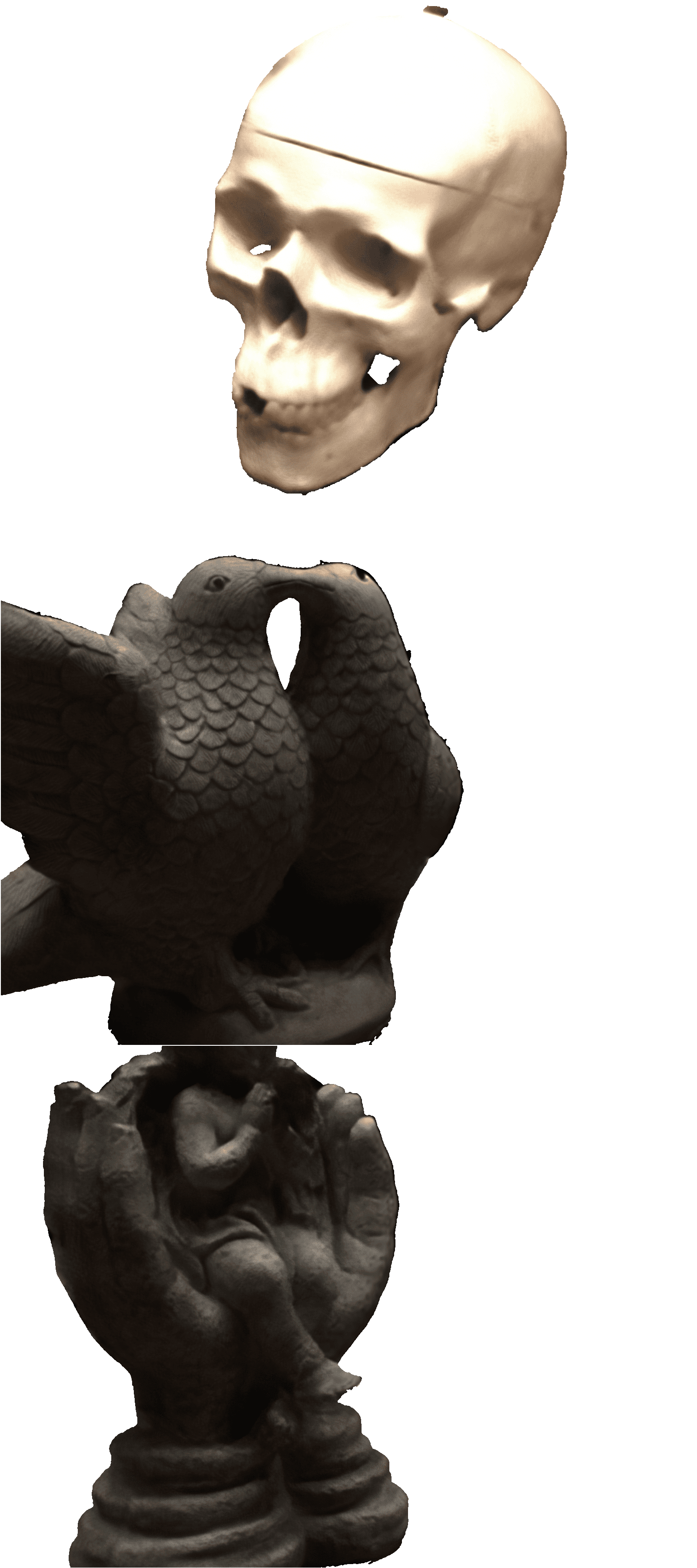}
         \captionsetup{font=tiny,labelfont=tiny}
        \caption{Ours (Single)}
     \end{subfigure}
     \begin{subfigure}[t]{0.22\linewidth}
         \centering
         \includegraphics[width=\linewidth]{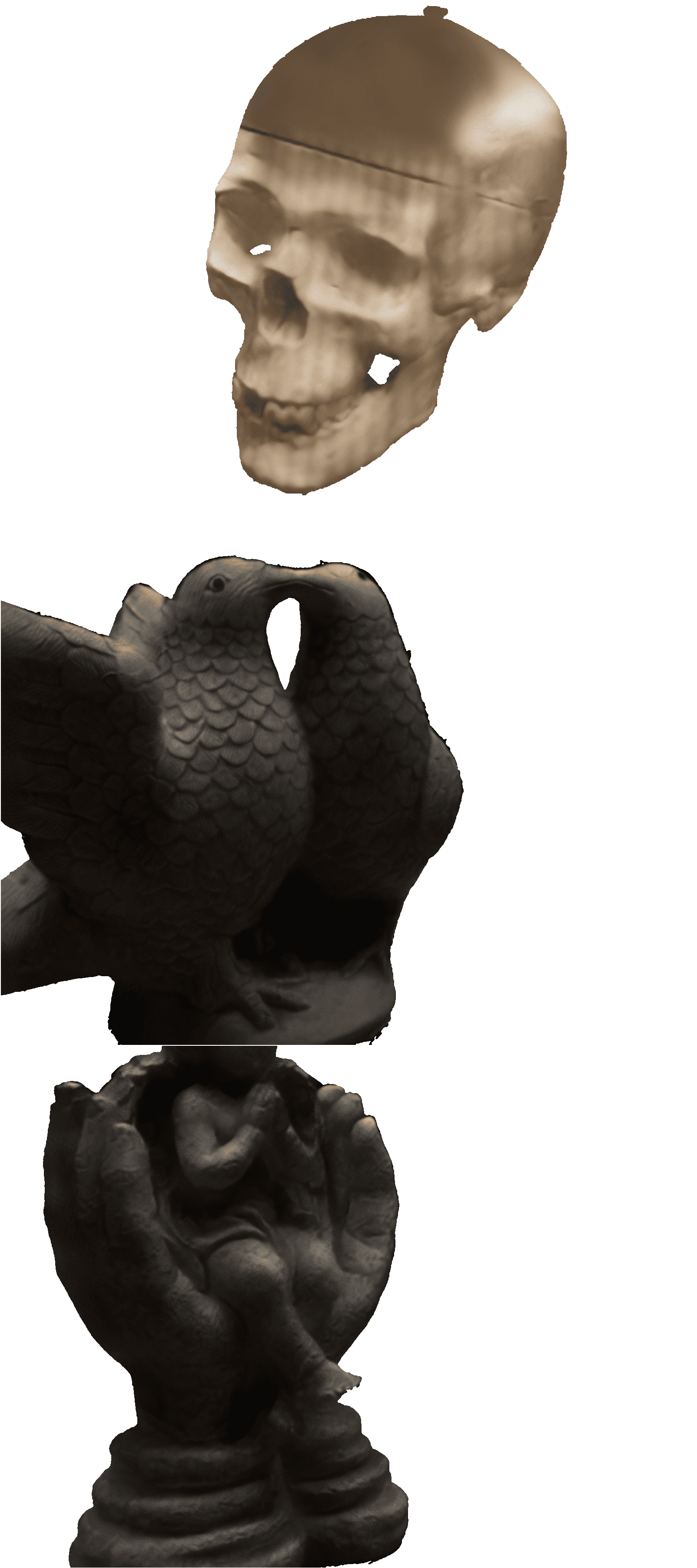}
         \captionsetup{font=tiny,labelfont=tiny}
        \caption{NPBG~\cite{aliev2020neural} (Full)}
     \end{subfigure}
     \begin{subfigure}[t]{0.22\linewidth}
         \centering
         \includegraphics[width=\linewidth]{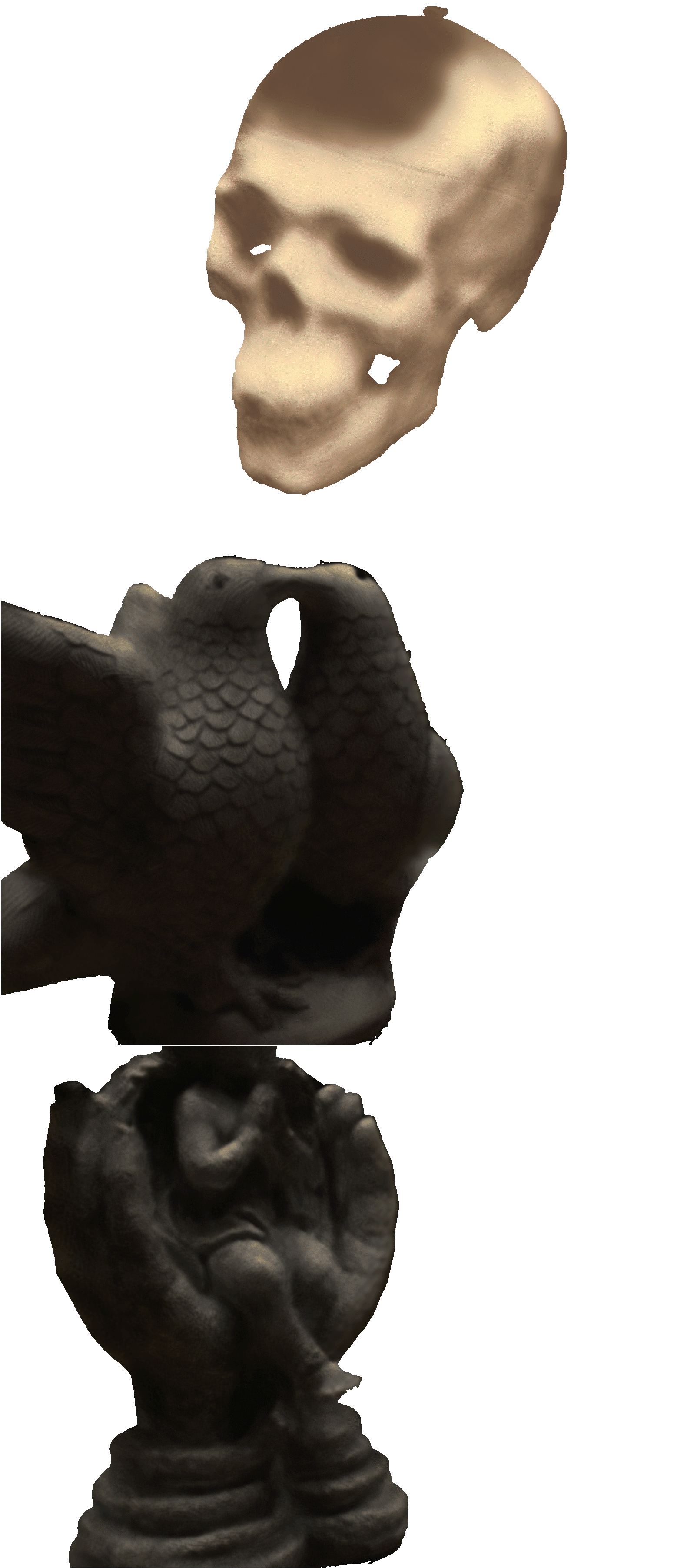}
         \captionsetup{font=tiny,labelfont=tiny}
        \caption{NPBG~\cite{aliev2020neural} (Single)}
     \end{subfigure}
 \caption{Additional qualitative comparisons on DTU~\cite{aanaes2016large}.}     
 \label{fig:qualitative_DTU_compare}
\end{figure*}

\paragraph{Comparison on Tanks and Temples~\cite{knapitsch2017tanks}:} Figure \ref{fig:qualitative_T_T_compare} shows additional novel view synthesis qualitative result comparison between our method for more test views on the Tanks and Temples~\cite{knapitsch2017tanks} dataset, not seen during training against SVS~\cite{riegler2021stable} and NeRF++~\cite{zhang2020nerf++}.
\paragraph{Comparison on DTU~\cite{aanaes2016large}:} Figure \ref{fig:qualitative_DTU_compare} shows additional novel view synthesis qualitative result comparison between our method for more test views on the DTU~\cite{aanaes2016large} dataset, not seen during training against SVS~\cite{riegler2021stable} and NeRF++~\cite{zhang2020nerf++}.

\clearpage
%
%
\bibliographystyle{splncs04}
\bibliography{egbib}